\documentclass[10pt,twocolumn,letterpaper]{article}

\usepackage{wacv}
\usepackage{times}
\usepackage{epsfig}
\usepackage{amsmath}
\usepackage{amssymb}
\usepackage{booktabs}
\usepackage{multirow}
\usepackage{graphicx}
\usepackage{xcolor} 
\usepackage{colortbl}
\usepackage{lipsum}  
\usepackage{parskip} 
\usepackage{makecell}
\usepackage{tabularx}
\usepackage{subcaption}
\usepackage{rotating}

\definecolor{lightgrayopacity}{gray}{0.9}

\def\wacvPaperID{****} %

\wacvapplicationstrack %

\wacvfinalcopy %

\ifwacvfinal
\usepackage[breaklinks=true,bookmarks=false]{hyperref}
\else
\usepackage[pagebackref=true,breaklinks=true,colorlinks,bookmarks=false]{hyperref}
\fi

\newcommand*{\affmark}[1][*]{\textsuperscript{#1}}
\newcommand*{\email}[1]{\texttt{#1}}
\begin{document}

\title{Noise Filtering Benchmark for Neuromorphic Satellites Observations}

\author{Sami Arja\affmark[*], Alexandre Marcireau, Nicholas Owen Ralph, Saeed Afshar and Gregory Cohen \\
Western Sydney University\\
{\email{$^{*}$s.elarja@westernsydney.edu.au}}
\and
}

\thispagestyle{empty}
\twocolumn[{
    \renewcommand\twocolumn[1][]{#1}
    \maketitle
    \includegraphics[width=\textwidth]{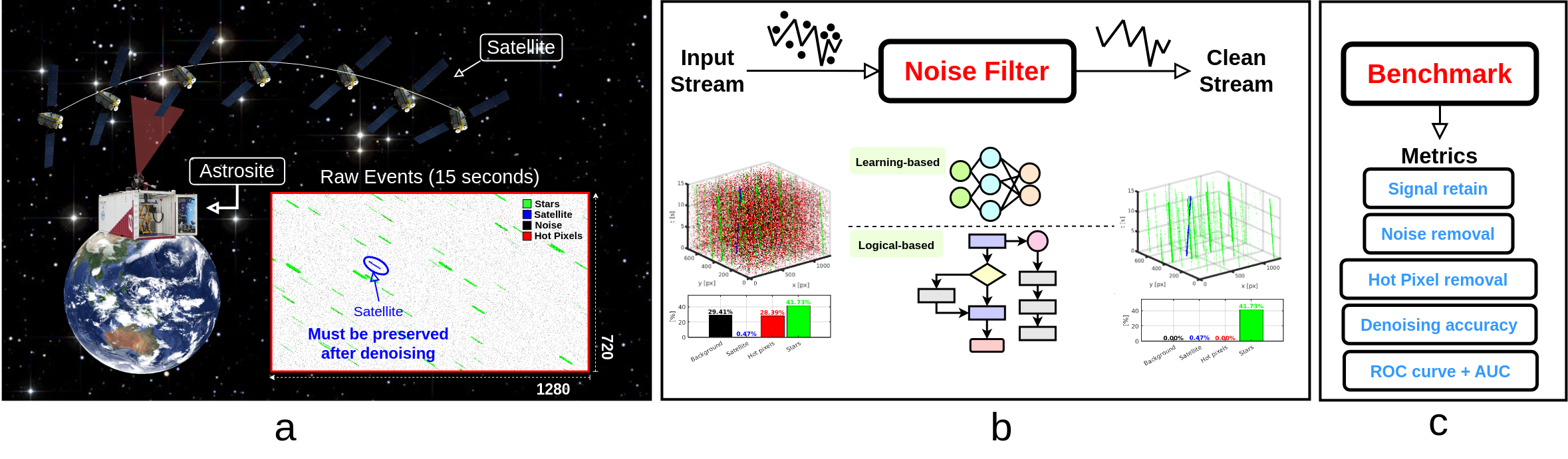}
    \captionof{figure}{The neuromorphic satellites observations application. \textbf{(a).} Satellite data is captured by a specialised mobile observatory (Astrosite) using an EVK4-HD event camera, showing an accumulated image over a 15-second recording with stars, background noise, and faint, glinting satellite signals. \textbf{(b)-(c).} The main goal of this paper is to benchmark all publicly available logic-based and learning-based algorithms for the noise filtering/removal task on sparse satellite data. The best-performing algorithm effectively denoises the sparse, noisy event stream while fully preserving the satellite signals. In addition, this paper proposes a new noise filtering algorithm, Cross-Convolution, and demonstrate the noise filtering capability of FEAST algorithm~\cite{afshar_event-based_2020} with and without a classifier.} \vspace{2em}
    \label{fig:long}
}]

\begin{abstract}
   Event cameras capture sparse, asynchronous brightness changes which offer high temporal resolution, high dynamic range, low power consumption, and sparse data output. These advantages make them ideal for Space Situational Awareness, particularly in detecting resident space objects moving within a telescope's field of view. However, the output from event cameras often includes substantial background activity noise, which is known to be more prevalent in low-light conditions. This noise can overwhelm the sparse events generated by satellite signals, making detection and tracking more challenging. Existing noise-filtering algorithms struggle in these scenarios because they are typically designed for denser scenes, where losing some signal is acceptable. This limitation hinders the application of event cameras in complex, real-world environments where signals are extremely sparse. In this paper, we propose new event-driven noise-filtering algorithms specifically designed for very sparse scenes. We categorise the algorithms into logical-based and learning-based approaches and benchmark their performance against 11 state-of-the-art noise-filtering algorithms, evaluating how effectively they remove noise and hot pixels while preserving the signal. Their performance was quantified by measuring signal retention and noise removal accuracy, with results reported using ROC curves across the parameter space. Additionally, we introduce a new high-resolution satellite dataset with ground truth from a real-world platform under various noise conditions, which we have made publicly available. Code, dataset, and trained weights are available at \url{https://github.com/samiarja/dvs_sparse_filter}.
\end{abstract}

\section{Introduction}

Event cameras, also known as Dynamic Vision Sensors (DVS) \cite{lichtsteiner_128times128_2008,Finateu2020510A1}, are optical sensors inspired by biological eyes. Unlike traditional cameras that capture the global brightness at regular intervals, DVS cameras detect changes in brightness at each pixel asynchronously, reporting these changes as log-intensity changes. These capabilities allow them to acquire visual information at the same rate as the scene dynamics, offering extremely high temporal resolution, low data rate output and a high dynamic range. Because of these features, DVS cameras can respond rapidly to dynamic scenes, making them ideal for satellite survey observations~\cite{afshar_event-based_ssa}.

The output of a Dynamic Vision Sensor (DVS) often contains a significant amount of undesirable background activity noise, which can blend with the actual signal and limit its effectiveness in practical applications. These background activity noise events originate from pixels even in the absence of any scene activity, rendering them noninformative. Background activity noise is primarily caused by photon and electron shot noise, as well as leakage in the reset transistor~\cite{graca_optimal_2023}. Notably, the noise rate increases dramatically under low-light conditions. The noise performance of a DVS depends not only on scene illumination but also on the camera's user-controllable bias settings. Therefore, adjusting the bias can potentially improve the clarity of the received signal~\cite{Graca_2023_CVPR,Mcreynolds2023ExploitingAD,nozaki_temperature_2017,graca_unraveling_2021}. However, large portions of the signal may still be compromised by noise from event jitter, quantization noise~\cite{Hu_2021_CVPR} and hot pixels (i.e. pixels with an abnormally high firing rate), which decrease the Signal-to-Noise Ratio (SNR) and increase the data rate~\cite{guo_low_2023,Hu_2021_CVPR}. Noise filtering the event stream at the camera output improves the SNR and reduces computational demands, thereby decreasing power consumption by orders of magnitude.

\begin{figure}[htbp]
\begin{center}
   \includegraphics[width=\linewidth]{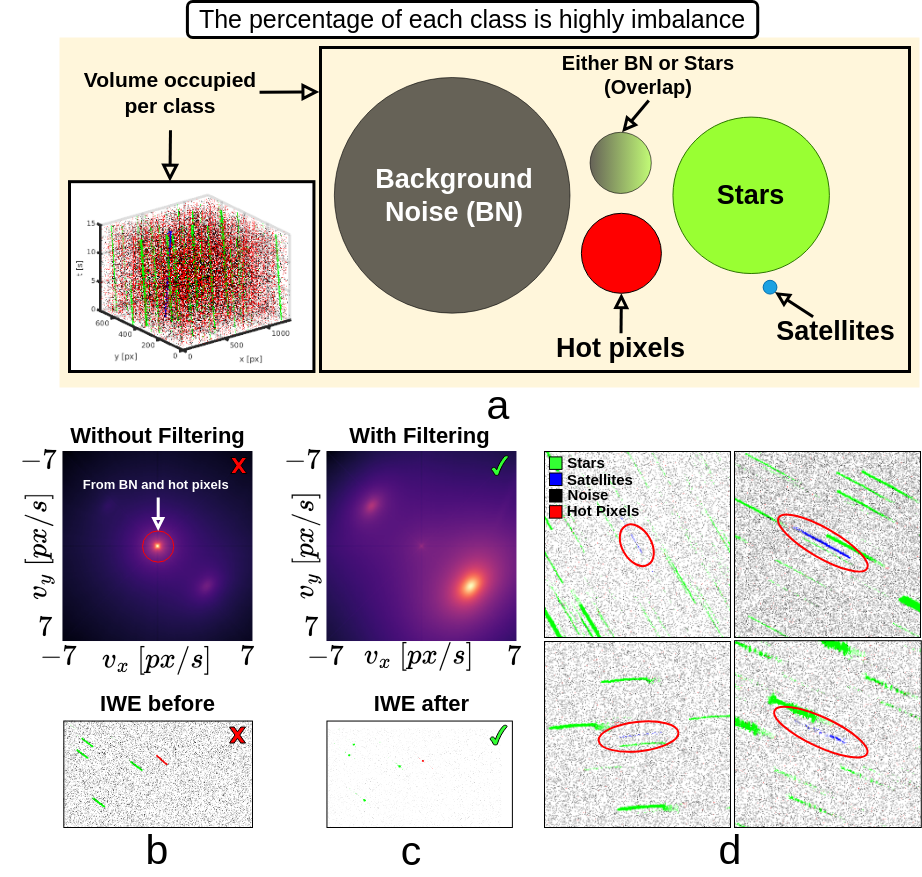}
\end{center}
   \caption{Overview about the satellite dataset from an event camera. (a) Event stream sizes for each class, where satellites generate the fewest events, while hot pixels and background noise dominate. Faint stars blend with background noise, making them hard to distinguish. (b) Motion landscape before noise filtering, showing high contrast at zero speed, resulting in a poor Image of Warped Events (IWE). (c) Motion landscape after noise filtering, revealing distinct camera and satellite motions, resulting in a sharp IWE. (d) Example of the satellite dataset highlighting their sparseness and the surrounding hot pixels and background noise.}
\label{fig:denoisingmotivation}
\end{figure}

To estimate the motion vector within the data, we utilised the Contrast Maximization (CMax) framework~\cite{gallego_unifying_2018}, which generates a motion landscape representing event activity across the velocity fields. In Figures~\ref{fig:denoisingmotivation}(b), the CMax-generated motion landscape reveals a peak at zero speed ($v_x = 0 \ [\text{px/s}]$, $v_y = 0 \ [\text{px/s}]$). This peak corresponds to unwanted background activity noise, indicating a concentration of stationary events in the image plane. An effective noise filtering algorithm should eliminate this peak by removing only the background activity noise, thereby enabling accurate motion compensation, as shown in Figure~\ref{fig:denoisingmotivation}(c).

In the broader neuromorphic and computer vision communities, eliminating unwanted noise to enhance visual data quality is referred to by various terms such as "denoising," "noise removal," "background activity filtering," and "noise filtering." In this paper, we use the term noise filtering to describe this process. Despite different terminology, the ultimate goal is the same: to identify and remove noisy events from the data stream. The key distinction among approaches lies in problem formulation. Some methods treat noise filtering as a binary classification task, requiring learning and large datasets; we refer to these as learning-based methods. Others simplify it to statistical calculations based on heuristics and specified conditions; we call these logical-based methods. In this paper, we formulate the noise filtering task as a multi-step statistical calculation for the CrossConv algorithm (i.e. logical-based) and as a binary classification task for the FEAST algorithm (i.e. learning-based).

This paper focuses on events produced by satellites which are the least dominant and the most valuable information for satellite surveying. Unlike stars, satellites are more trivial to label since they are usually isolated and move in the opposite direction from the telescope. Additionally, determining the limiting magnitude of star brightness from an event camera remains an open research question~\cite{marcireau2023binocular,ralph_astrometric_2023,mcreynolds_demystifying_2023}, as faint stars often blend with noise, making them difficult to identify and label. For this reason, our evaluation of the noise-filtering algorithms is based solely on how well they preserve satellite events. In this paper, we investigate whether noise filtering is ideal for very sparse satellite data and, if so, whether logical-based or learning-based approaches should be employed. We define logical-based algorithms as those that use deterministic rules to distinguish signal events from noise events. In contrast, learning-based algorithms learn features of the signal and noise during a training phase and perform noise filtering during inference.

Prior noise filtering algorithms include bioinspired filter~\cite{barrios2018less}, hardware-based filter~\cite{khodamoradi_on-space_2018,liu2015design}, spatial filter~\cite{benosman2013event,delbruck_frame-free_2008} and temporal filter~\cite{baldwin_edncnn,karray_inceptive_2019,wang2020joint,fang_aednet_2022}, aiming to achieve real-time processing. While effective in some scenarios, they often struggle with the low SNR in data from small, sparse objects like satellites. These methods may inadvertently remove signal events along with noise, which is undesirable for precise applications such as satellite tracking.

To ensure a fair comparison between noise filtering algorithms, we first established a high-quality ground truth and employed the evaluation strategy from~\cite{guo_low_2023}, using Receiver Operating Characteristic (ROC) curves across a wide parameter space for each algorithm. Additionally, we utilized evaluation metrics such as Signal Retain (SR), Noise Removal (NR), and Denoise Accuracy (DA) from~\cite{duan_led_2024}, and introduced a new metric, Hot Pixel Removal (HPR). We also evaluated how long a recording needs to be for each algorithm to achieve maximum noise-filtering performance. Knowing the minimal required event duration is beneficial, as it allows for faster processing and optimal use of resources. Furthermore, we propose a new lightweight logical-based algorithm called "CrossConv" designed to efficiently eliminate both low and high-activity hot pixels and background noise. We also evaluated the noise filtering capabilities of the learning-based FEAST~\cite{afshar_event-based_2020} algorithm (Feature Extraction with Adaptive Selection Thresholds).

The contribution of this work can be summarised as follows:

\begin{itemize}
    \item We constructed an event-based dataset of satellites called "Ev-Satellites" with high-quality ground truth to enable precise noise filtering evaluation.
    \item We proposed a novel logical-based algorithm called CrossConv and evaluated the performance of the learning-based FEAST~\cite{afshar_event-based_2020} algorithm, which had not been previously assessed for noise-filtering tasks. 
    \item We conducted an extensive comparison against 11 state-of-the-art noise-filtering methods, rigorously quantifying their performance using metrics such as Signal Retain (SR), Noise Removal (NR), Denoise Accuracy (DA), Hot Pixel Removal (HPR) and Receiver Operating Characteristic (ROC) curve.
\end{itemize}

\section{Related Work}

\textbf{DVS Noise Characteristics}. DVS pixels often register events even when there's no intensity change, primarily due to thermal and shot noise~\cite{graca_unraveling_2021}, often referred to as background activity. background activity arises from two main sources: photon/electron shot noise and leakage current~\cite{graca_unraveling_2021,nozaki_temperature_2017}. Photon/electron shot noise is the dominant noise source in low-brightness environments, while leakage current noise becomes more prominent in high-brightness conditions~\cite{graca_unraveling_2021,nozaki_temperature_2017}. The sensor's bias settings can also contribute to background activity which can be controlled to some extent by tuning them either manually or automatically~\cite{Delbruck_2021_CVPR}. Another type of noise comes from "hot pixels," which generate a high rate of noise events due to transistor and device mismatches~\cite{guo_low_2023}. In low-light conditions, such as under a night sky, the frequency of these noise events can significantly increase. For satellite surveillance, distinguishing the noise from the signal is challenging, as sparse, small satellites result in extremely low SNR.

\textbf{Noise Filtering Datasets and Metrics}. Several task-specific datasets have been introduced to the event-based community to evaluate the performance of various noise filtering algorithms. These datasets differ in terms of the number of sequences, duration, whether they are simulated or based on real-world scenes, noise levels, illumination rates, and resolution. For example, DVSCLEAN~\cite{fang_aednet_2022} and DND21~\cite{guo_low_2023} were generated using event camera simulators. In contrast, datasets like DVSNOISE20~\cite{baldwin_edncnn} were captured from real-world scenes with both stationary and moving cameras, as well as stationary cameras with moving objects. RGB-DAVIS~\cite{wang2020joint} includes 20 indoor and outdoor sequences recorded under good lighting conditions, while ED-KoGTL~\cite{alkendi_neuromorphic_2024} provides indoor data with four different illumination levels under a constant trajectory. E-MLB~\cite{ding_e-mlb_2024} offers a large dataset with varying noise levels under both low and high illuminations. LED~\cite{duan_led_2024} presents a high-resolution dataset, the largest to date in terms of the number of sequences and sequence duration. However, these datasets are primarily designed for dense scenes, making it difficult to evaluate algorithm performance on very sparse signals. To address this, we introduce a new dataset, "Ev-Satellites," which is sparse and features varying brightness levels, containing an arbitrary number of stars and satellites with different trajectories.

Various metrics have been proposed to more accurately quantify noise filtering performance. These include the percentage of signal/noise remaining (PSR/PNR)~\cite{padala_noise_2018}, which is based on a bounding box, and the relative plausibility measure of noise filtering (RPMD)~\cite{baldwin_edncnn}, which combines APS and IMU data to calculate the probability of event occurrence at each space-time coordinate. Event Denoising Precision (EDP)~\cite{wu_probabilistic_2021} measures the ratio of the total number of denoised events to the original event stream. Guo et al. \cite{guo_low_2023} reported algorithm performance across the entire parameter space using the Receiver Operating Characteristic (ROC) curve, which plots the True Positive Rate (TPR) against the False Positive Rate (FPR) across varying discrimination thresholds. \cite{duan_led_2024} introduced metrics like Signal Retain (SR), Noise Removal (NR), and Denoise Accuracy (DA) to assess how well an algorithm preserves signal versus how effectively it removes background activity. In this work, we evaluate state-of-the-art noise filtering algorithms across the entire parameter space to ensure a fair comparison based on the ROC curve~\cite{guo_low_2023}. We also use SR, NR, and DA as performance metrics based on~\cite{duan_led_2024} and introduce a new metric, Hot Pixel Removal (HPR), to further enhance our evaluation.

\textbf{Noise Filtering Algorithms}. Noise filtering algorithms generally fall into two categories: logic-based and learning-based. The former~\cite{guo_low_2023,feng_event_2020,lagorce_hots_2017,wang_ev-gait_2019,karray_inceptive_2019,khodamoradi_on-space_2018,delbruck_frame-free_2008,guo2020hashheat,acharya2019ebbiot,mohan2020ebbinnot,cladera2020device,bose202151} uses manually crafted rules and conditions to distinguish between noise and signal, while the latter~\cite{guo_low_2023,fang_aednet_2022,afshar_event-based_2020,Duan_2021_CVPR,baldwin_edncnn,linares2019low} leverages deep neural networks to learn the common patterns in noise and signal. Some algorithms in both categories are also optimised for hardware implementation~\cite{delbruck_frame-free_2008,afshar_event-based_2020,khodamoradi_on-space_2018,guo2020hashheat,acharya2019ebbiot,mohan2020ebbinnot,cladera2020device,linares2019low}. Logical-based algorithms do not require a prior about the scene and can be applied to a wide range of scenarios, but their accuracy is constrained by the thresholds and rules they employ. On the other hand, learning-based algorithms tend to achieve higher accuracy by adapting to specific data patterns. However, they are typically limited to the datasets on which they were trained and must be retrained for new tasks. 

In this paper, we benchmark the publicly available noise filtering algorithms~\cite{guo_low_2023,feng_event_2020,lagorce_hots_2017,wang_ev-gait_2019,karray_inceptive_2019,khodamoradi_on-space_2018,fang_aednet_2022,afshar_event-based_2020} that process event-only input. While other noise filtering methods exist in the literature, such as EDnCNN~\cite{baldwin_edncnn} and GEF~\cite{duan_guided_2021}, these algorithms incorporate additional data modalities beyond events. EDnCNN trains a classification network using probability labels for each event, estimated by combining information from Active Pixel Sensors (APS) and Inertial Measurement Units (IMU). GEF removes inconsistent events by extracting mutual structures between the event frame and the gradient of the APS image. Since our focus is on algorithms that operate solely on event data, we have limited our evaluation to methods that meet this criterion.

\textbf{Space Situational Awareness (SSA)}. Given the inherent high time resolution, low latency, and low volume data output of the DVS, it has shown promise for SSA applications such as space imaging and Resident Space Object (RSO) tracking~\cite{cohen2018approaches,cohen2019event,mcmahon2021commercial,zolnowski2019observational}. DVS provides fast, accurate and sufficient amount of information of RSOs with minimal integration time and with high response rate. 

Motivated by the capabilities of DVS in SSA applications, the Astrosite a containerized event-based telescopic space imaging system~\cite{cohen2018approaches} was built. Its viability has been successfully demonstrated in star imaging~\cite{ralph2023astrometric,ralph2023shake, marcireau2023binocular} and in RSOs tracking, such as satellites~\cite{afshar_event-based_ssa,ralph2022real, oliver2022event,ralph2023exploring}.
Previous work~\cite{mcreynolds_demystifying_2023} pushed the performance limits of the DVA and investigated the effect of DVS bias for capturing sparse satellite information. \cite{jolley2023neuromorphic} performed satellite characterisation using DVS event-rate output to determine the rotation rate of nonstabilized satellites. \cite{jolley2022evaluation,jolley2022characterising,jolley2019use} demonstrated the use of the DVS for satellite material characterisation based on their broadband reflectance properties. In this thesis, \cite{Oliver2024} proposed a physics-based end-to-end model to simulate event data for SSA applications. In this work, we focus on methods and techniques to improve the quality of the DVS output for RSO observations and improved space imaging via event-based noise filtering.

\section{Ev-Satellites Data}

\subsection{Preliminaries}

The DVS is a 2D matrix of pixels that works asynchronously and will trigger an event $e_i=(\bold{u_i},p_i,t_i)$ as a 4-tuple when its logarithmic brightness change reaches the predefined contrast threshold. In this tuple, $\bold{u_i}=(x_i,y_i)$ is the pixel address of the $i\_th$ event, $p_i=\pm 1$ is the polarity of the brightness change, indicating ON or OFF events and $t_i$ is the event's timestamp in microseconds.

Since DVS cameras detect changes in brightness, they are sensitive to the moving edges of objects. As long as there is relative motion between the camera and the objects, a series of events will occur along the edge trajectories. To create a star or satellite map, we typically scan (i.e. translate) the sky over a period of time $\Delta t$ to generate contrast. During this period, we count the number of events generated by each pixel to form an accumulated frame, which shares many characteristics with conventional astronomical images. However, because stars and satellites appear as streaks due to the motion, we need to compensate for this motion using the CMax framework~\cite{gallego_unifying_2018} to transform them into point sources. It works by taking a set of events $\varepsilon = \left\{ e_i \right\}^{N_e}_{i=0}$ and aligning them with the edges or objects that generated them, by estimating the camera's relative motion vector $\boldsymbol{\theta}=[v_x,v_y]$ over a time window $\delta=t_i - t_{ref}$. This operation is equivalent to a shear transformation (i.e. warping) of the event point cloud $\varepsilon$ which reverses the motion $\boldsymbol{\theta}$ between $t_i$ and the beginning of $\delta$ and changes the spatial location of the events $\boldsymbol{u_i}$, described as $\boldsymbol{u'_i} = \boldsymbol{u_i}- \bold{\theta} * \delta$. The events are accumulated into an Image of the Warped Events (IWE) by summing $H(\boldsymbol{u'};\boldsymbol{\theta}) \dot{=}\sum_{i=1}^{N}b_{k}\delta(\boldsymbol{u}-\boldsymbol{u'_{i}})$. The contrast is then calculated as a function of $\theta$ using the variance as the objective function~\cite{gallego_unifying_2018,gallego2019focus}, described as $\sigma^2(H(\boldsymbol{u'};\boldsymbol{\theta}))$. The goal of CMax is to identify the correct $\boldsymbol{\theta}$ that maximizes the variance which can be performed using an optimisation algorithm.

\subsection{Dataset Collection Details}
\begin{figure}[h]
\begin{center}
   \includegraphics[width=0.9\linewidth]{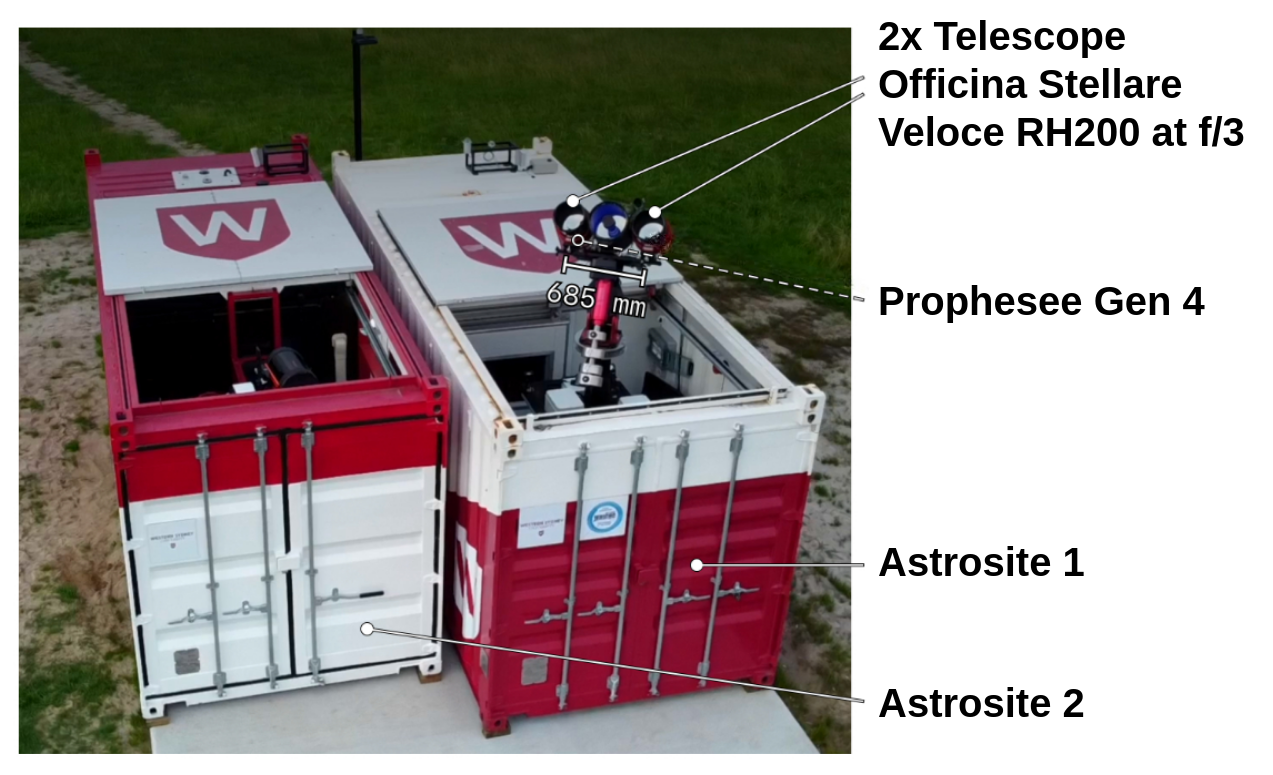}
\end{center}
   \caption{Astrosite setup. The data was collected with Astrosite 2 deployed in Regional South Australia (Adelaide). Image adapted from~\cite{marcireau2023binocular}.}
\label{fig:astrosite_setup}
\end{figure}

The data was collected in this paper using a specialised observatory built specifically for Event-based Space Situational Awareness (EBSSA) called the "Astrosite". It is a 20ft standard shipping container with a sliding roof, and a scissor lift that supports a motorised mount with three bore sighted telescope~\cite{cohen2018approaches,ralph2023exploring,ralph_astrometric_2023,marcireau2023binocular} (Figure~\ref{fig:astrosite_setup}). It is equipped with a Gen 4 HD sensor designed by Prophesee and Sony (1280 $\times$ 720 pixels in a 1/2.5" format, with a pixel size of 4.86 $\mu$m × 4.86$\mu$m) mounted on a Planewave L600 Altazimuth telescope with a field of view of 35.6' × 20.0' with a pixel scale of 1.669 arcsec/px. Astrosite has been fully automated as of 2024, enabling it to self-deploy from sunset to sunrise, depending on weather conditions. Each night, it captures data from an average of 800-1000 satellites. The event rate and noise levels in the dataset vary due to factors such as sky brightness, cloud cover, sensor bias, and weather conditions. To ensure a diverse dataset, we selected 100 event files recorded between June 2022 and June 2024, selecting those with higher noise levels. Each recording, lasting 30 seconds, includes events related to stars of varying brightness and was obtained by slewing the telescope at a constant speed through the sky. In this setup, static stars appear to move in the opposite direction of the camera motion, while satellites move in the opposite direction of the stars. The data is stored in the event stream format\footnote{https://github.com/neuromorphicsystems/event\_stream}. Figure~\ref{fig:denoisingmotivation}(e) provides an example of how the satellites appear, surrounded by noise, hot pixels, and other star events (in green). We refer to this dataset as "Ev-Satellites".

\subsection{Framework for Generating Ground Truth}

\begin{figure}[h]
\begin{center}
   \includegraphics[width=0.9\linewidth]{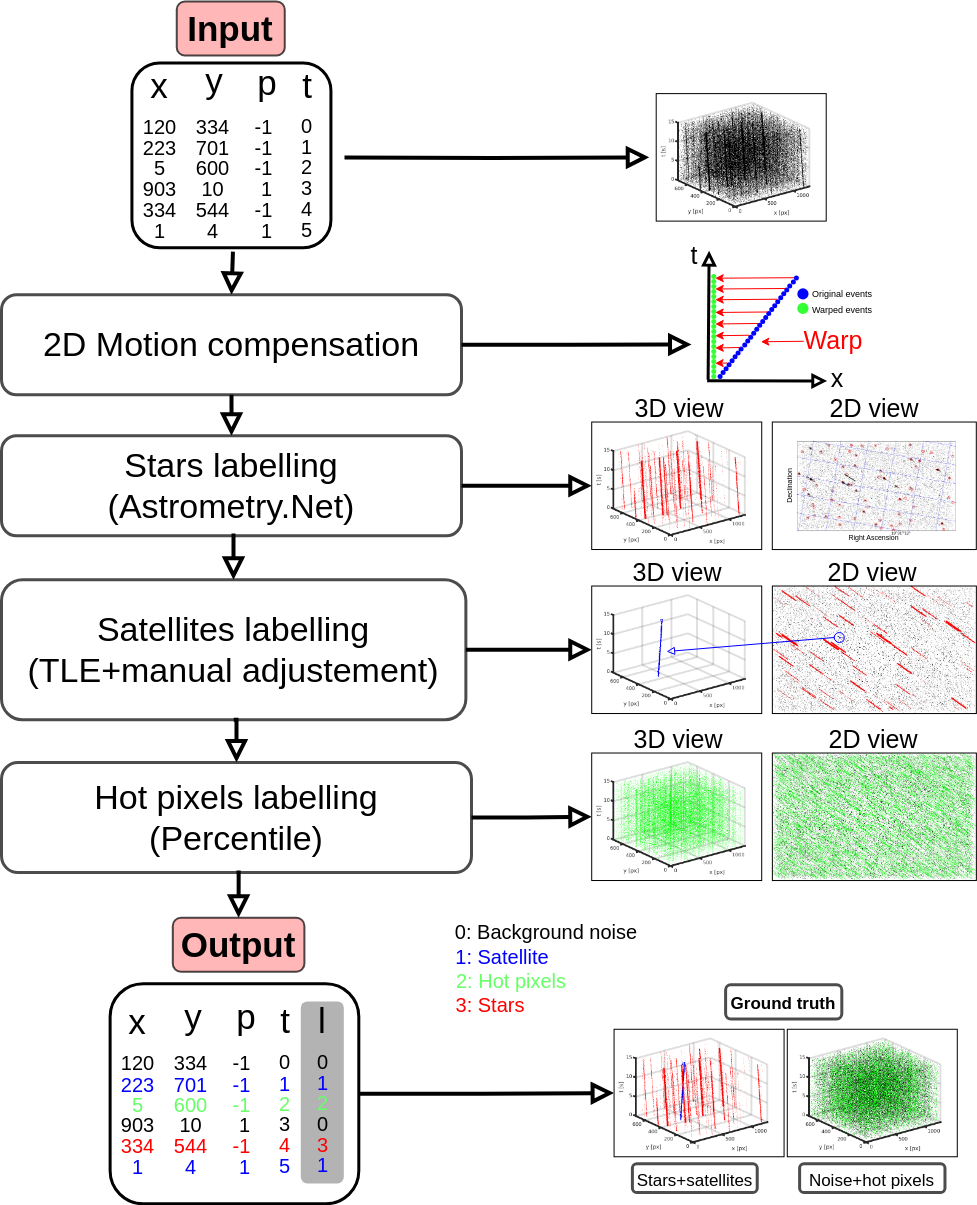}
\end{center}
   \caption{Labelling process for the "Ev-Satellites" dataset, detailing the steps required to categorize each event as either a star, satellite, hot pixel, or noise. The goal is to augment the events by returning the per-event label array "l".}
\label{fig:labellingscheme}
\end{figure}

Our proposed framework for event labelling begins with stars, followed by satellites, and then hot pixels, as shown in Figure~\ref{fig:labellingscheme}. The process starts by estimating the field velocity $\boldsymbol{\theta}$ with CMax. Due to the presence of excessive noise and hot pixels, we manually initialized the $\boldsymbol{\theta}$ near its actual value to ensure faster and more accurate convergence. Without this adjustment, the process could converge to incorrect maxima, as demonstrated in Figure~\ref{fig:denoisingmotivation}(c)-(d).

Once the field velocity is estimated, it is used to generate a motion-compensated image $H(\boldsymbol{u'};\boldsymbol{\theta})$, which allows for the creation of star maps from the event data. A connected components analysis algorithm (CCA) is employed to cluster the stars and calculate their centres and sizes, using a sensitivity parameter defined as $\rho=1-\frac{k}{100}$.

The list of star centers from the CCA algorithm, along with the known mount position, is used as input to the Astrometry.net algorithm\footnote{https://github.com/neuromorphicsystems/astrometry}~\cite{lang2010astrometry}, which retrieves star locations from the Gaia DR3 catalog~\cite{andrae2023gaia}. Previous research~\cite{ralph_astrometric_2023,marcireau2023binocular} has established the limiting magnitude to be between 14 and 15. To push the limiting magnitude to the far end of 14, we used a highly sensitive parameter $\rho=0.9$ to capture as many stars as possible with a non-zero event count.

One of the challenges is differentiating between very faint stars and noisy pixels, particularly at higher limiting magnitudes, due to the high noise levels. As a result, some very faint stars may remain unlabelled. However, their percentage is minimal, and thus their impact on our satellite evaluations is negligible.

Astrometry.net provides the positions of the stars (x,y) along with their magnitudes. Since magnitude is inversely proportional to brightness, we adjusted the size of the circles around the stars to also be inversely proportional to brightness. All pixels within each circle, along with their associated events, were labelled as stars (label=3). These labelled events were then removed from the event stream, leaving only the remaining satellites and background noise.

Each recording is accompanied by Two Line Elements (TLE) data, which provides the satellite's location in space~\cite{vallado2012two}. However, TLE data can be imprecise and may not align perfectly with the target satellite. Developing an accurate orbit propagation algorithm is beyond the scope of this paper. Therefore, we used the TLE data as an initial estimate of the satellite's location and then estimated its velocity to produce a motion-compensated image. We manually placed circles around the satellite in the image and labelled all the pixels within these circles as satellite (label=1). These labelled satellite events were then removed from the event stream, leaving only the remaining background noise with the hot pixels.

Hot pixels exhibit a higher firing rate compared to background noise. To label them, we treat them as outliers by computing the 98th percentile of the image, which includes background noise, and remove the top 2\% of the brightest values. The events corresponding to these brightest pixels are labelled as hot pixels (label=2). The remaining events are labelled as background noise (label=0). This labelling process produces a new array with the same size as the original event data, where each row specifies the label for each event which results in 5-tuple $e_i=(\bold{u_i},p_i,t_i,label_i)$.

\begin{figure*}[htbp]
\begin{center}
   \includegraphics[width=0.9\linewidth]{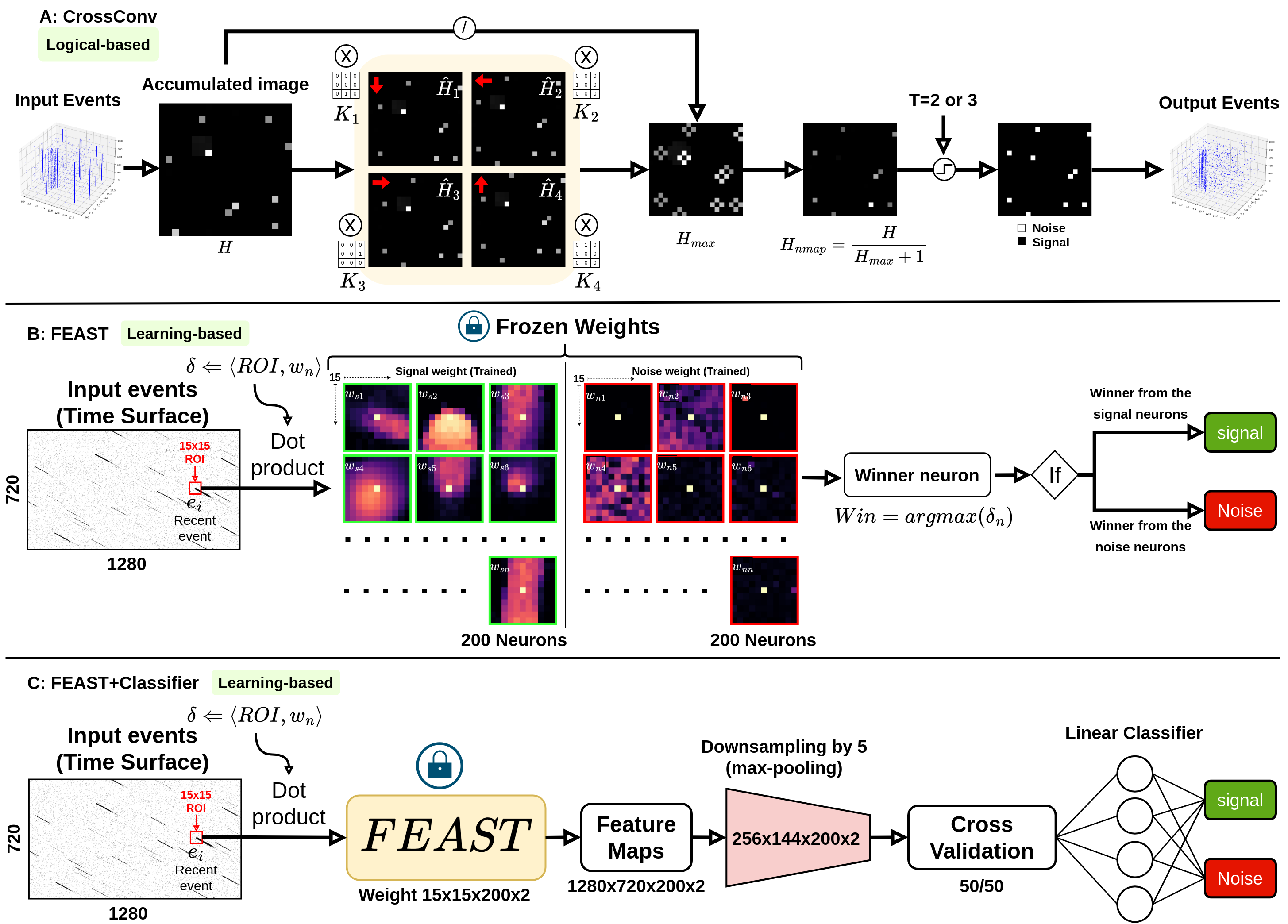}
\end{center}
   \caption{Overview of the noise filtering algorithms used in this paper. (A) The pipeline of our proposed CrossConv noise filtering algorithm is categorized as a logic-based approach. It processes the event point cloud in the image space to identify high-firing rate pixels and then returns to the event space to remove them. (B)-(C) The FEAST algorithm~\cite{afshar_event-based_2020}, which classifies events directly during inference based on the winner neuron for each class after being trained in a supervised manner (B), and the same architecture using a linear classifier (C).}
\label{fig:algorithm_architectures}
\end{figure*}

\section{Noise filtering Algorithms}

Figure~\ref{fig:algorithm_architectures} illustrates the proposed noise filtering algorithms in this paper which we compared with other publicly available noise filtering methods. Below we describe our algorithms.

\textbf{Cross-Convolution}.  We refer to it as "CrossConv". This algorithm filters out pixels that have a higher firing rate even with the presence of other objects in motion. Overall, it prioritises removing all kinds of hot pixels (e.g. short and long-lasting ones) and it leaves noise pixels that have a very sparse firing rate. As illustrated in Figure~\ref{fig:algorithm_architectures}(A), CrossConv algorithm takes a set of events $\varepsilon$ and accumulate them into an image $H(\boldsymbol{u'};\boldsymbol{\theta})$ with $\boldsymbol{\theta}=[0,0] \ px/[s]$. We convolve $H$ with 4 convolution kernels $K_i$, $\hat{H} = H\bigotimes K_i$
, where $\bigotimes$ denotes convolution. Each kernel is a \(3 \times 3\) matrix that shifts $H$ by one pixel down using $K_1=$\(\left(\begin{smallmatrix} 
\scriptstyle 0 & \scriptstyle 0 & \scriptstyle 0 \\ 
\scriptstyle 0 & \scriptstyle 0 & \scriptstyle 0 \\ 
\scriptstyle 0 & \scriptstyle 1 & \scriptstyle 0 
\end{smallmatrix}\right)\), left using $K_2=$\(\left(\begin{smallmatrix} 
\scriptstyle 0 & \scriptstyle 0 & \scriptstyle 0 \\ 
\scriptstyle 1 & \scriptstyle 0 & \scriptstyle 0 \\ 
\scriptstyle 0 & \scriptstyle 0 & \scriptstyle 0 
\end{smallmatrix}\right)\), right using $K_3=$\(\left(\begin{smallmatrix} 
\scriptstyle 0 & \scriptstyle 0 & \scriptstyle 0 \\ 
\scriptstyle 0 & \scriptstyle 0 & \scriptstyle 1 \\ 
\scriptstyle 0 & \scriptstyle 0 & \scriptstyle 0 
\end{smallmatrix}\right)\) and up using $K_4=$\(\left(\begin{smallmatrix} 
\scriptstyle 0 & \scriptstyle 1 & \scriptstyle 0 \\ 
\scriptstyle 0 & \scriptstyle 0 & \scriptstyle 0 \\ 
\scriptstyle 0 & \scriptstyle 0 & \scriptstyle 0 
\end{smallmatrix}\right)\). This operation produces a set of convolved images $\hat{H}=[\hat{H}_{k_1},\hat{H}_{k_2},\hat{H}_{k_3},\hat{H}_{k_4}]$. The following step is similar to a pixel-wise max-pooling operation, where for every pixel in $\hat{H}$ we select the pixel with the maximum intensity and add it to a new empty image. This process results in another image denoted as $H_{max}$ where the pixel with a high firing rate forms a plus-shape pattern. This operation creates another 3 pixels for each pixel with firing rate as a form of pixel augmentation. After this, we divide $H$ with $H_{max}$ to create $H_{nmap}$ of the image of the noise map (nmap). This operation maximises the variance between the pixels triggered by a signal and by noisy pixels. Finally, a threshold is applied to $H_{nmap}$ to create a binary image, where white pixels represent noise, and the remaining pixels are classified as signals. This threshold is the only parameter needed for this algorithm. We then remove all the events associated with each of the noisy pixels. Note that CrossConv will leave a few of the low firing rate noisy pixels, as it prioritises high firing rate pixels.

\textbf{FEAST}. \cite{afshar_event-based_2020} introduced the FEAST algorithm that extracts features from event-based data in an unsupervised manner using spiking neuron-like units
which have individual selection thresholds (as shown in Figure~\ref{fig:algorithm_architectures}(B)). Building on this, \cite{bethi_optimized_2022} proposed a multilayer version of FEAST that learns features in a supervised way and classifies event data without using gradient descent. In our approach, we use the original single-layer FEAST algorithm but restructure the initial learning phase into a supervised format by allowing it to learn features of stars, satellites, and noise. To our knowledge, we are the first to evaluate FEAST capabilities for event-based noise filtering task~\cite{guo_low_2023,afshar_event-based_2020}.

The FEAST algorithm uses the exponentially decaying time surface kernel~\cite{tapson2013synthesis,lagorce_hots_2017} that maps a time $t$ to each 2D spatial coordinate $\boldsymbol{u}$. The time surface is updated only when an event arrives at a channel and the weight of the information carried by each event decays smoothly toward zero over time. To do this, we define two distinct functions: $T_{i} \in \mathbb{R}^{H\times W}$ which maps a pixel position to a timestamp, expressed as $\boldsymbol{u} : t \mapsto T_i(\boldsymbol{u})$, and $P_{i} \in \mathbb{R}^{H\times W}$
which represents the polarity of the most recent event generated by pixel $u$, expressed as $\boldsymbol{u} : p \mapsto P_i(\boldsymbol{u})$. The process described in Eq.~\ref{eq:timesurface} encapsulates both the temporal dynamics and polarity information at each point in the observed space.

\begin{equation}
    I_i(\boldsymbol{u}, t) = P_i(\boldsymbol{u}) \cdot e^{\left(\frac{T_i(\boldsymbol{u})-t}{\tau}\right)} \label{eq:timesurface}
\end{equation}

where $\tau$ is the time constant parameter. An event context $C$ (i.e. Region of Interest) of size $r\times r$ is created around each incoming event containing spatiotemporal information from the neighbourhood surrounding the pixel. This limits the feature information to the local spatial context around the event and allows computationally efficient event-based processing. The event context is then vectorized and normalized through a division by its norm to achieve invariance to temporal scaling, $C_{norm} = \frac{C}{||C||}$. An $N$ neuron matrix $W_N \in \mathbb{R}^{r\times r}$ representing the synaptic weights are initialized randomly to learn the commonly occurring features in $C$. Each neuron has a threshold $\theta_N$ as an additional parameter that is also initialized randomly. For every incoming event, a dot product of the event context $C$ and each of the weights matrices $W$ is calculated as in, $d_N = C \cdot W_N$. The dot products of all neurons are then compared to their corresponding threshold values and the neuron with the largest dot products and with a value greater than their thresholds ($d_N \ge \theta_N$) is considered the winning neuron described as $\psi$, calculated like this $\psi=argmax(d_N)$. A winner neuron indicates that the information in the event context matches this neuron. The weight vector of this neuron is updated as follows,

\begin{equation}
   w_N = (1-\eta) w_N + \eta C\label{eq:weightupdate}
\end{equation}

Where $\eta$ is the learning rate. During training, the threshold for the winning neuron is increased by a fixed value, while the thresholds and weights of other neurons remain unchanged. If no matching neuron is found, the thresholds of all neurons are reduced by a predefined value. This dynamic threshold adaptation helps maintain homeostasis and ensures that feature neurons are activated equally in response to incoming events.

The FEAST algorithm, in its various forms, has been successfully applied to tasks like star tracking~\cite{ralph2023exploring}, fruit detection~\cite{el2022neuromorphic}, and supervised spike-based classification~\cite{bethi_optimized_2022}. However, in its standard implementation, FEAST tends to favour the dominant class, leading to imbalanced features skewed toward the class with more events. This is not ideal for noise filtering tasks because the distributions of the signal and noise are not balanced.

To address this, we implement supervised training by initializing two sets of weight templates $W_{N_{s}}$ and $W_{N_{n}}$, and two threshold matrices, $\theta_{N_{s}}$ and $\theta_{N_{n}}$, for signal and noise respectively, with the same number of neurons in each set. During training, weights and thresholds are updated based on the label of incoming events. Once training is complete, the weight matrices for signal and noise are saved for use during inference, and the threshold adaptation rules are discarded. The supervised FEAST can either be trained separately for each class or simultaneously by implementing a condition that checks the label of each incoming event.

At inference, to classify an event as either noise or signal, we calculate the dot product between the new event context $C_new$ from the test data and the trained weights $W_{N_{s}}$ and $W_{N_{n}}$ and if the winning neuron belongs to any feature from $W_{N_{s}}$ the event is then classified as a signal; otherwise, it is classified as noise. The whole inference process is performed in a single feed-forward pass for each incoming event.

\textbf{FEAST+Classifier}. We test FEAST noise filtering capabilities by adding a linear classifier at the output using the same trained features/weights for both classes as shown in Figure~\ref{fig:algorithm_architectures}(C). In this case, the winning neuron is no longer used to classify the incoming events, but rather a classifier that maps the input spatio-temporal spike pattern $X$ to any output spatio-temporal spike pattern $Y$ using a weight matrix $W_c$ with a defined number of neurons.

Similarly, for each incoming event, we generate an event context centred around it. We then compute the dot product \( d_N \) between this event context and each neuron in the trained weight matrices \( W_{N_s} \) and \( W_{N_n} \) to identify the winning neuron \( \psi \). We construct a three-dimensional feature map (or activation map) \( F \in \mathbb{R}^{H \times W \times N} \), where \( H \) and \( W \) are the spatial dimensions of the input, and \( N \) is the total number of neurons.

This feature map projects the activity of each neuron onto a two-dimensional spatial plane, effectively mapping events to their corresponding neurons. Specifically, if an event at position \( \boldsymbol{u_i} = (x_i, y_i) \) has neuron \( k \) as its winning neuron, we project $\boldsymbol{u_i}$ to the \( k \) feature map. This means that the event \( \boldsymbol{u_i} \) is associated with neuron \( k \) in the feature map at the spatial coordinates \( (x_i, y_i) \). Processing the entire feature map can be computationally intensive and memory-demanding due to its high dimensionality (e.g. $1280 \times 720 \times N \times E$, $E$ is the number of events). To address this issue, we apply spatial event-based downsampling to reduce the dimensions of the feature map. Instead of uniformly downsampling the entire feature map, we selectively downsample only the active regions—areas where events occur—while leaving the empty regions unchanged. This approach reduces computational load without compromising the classifier's performance, as demonstrated in earlier work~\cite{cohen2018spatial}. In this paper, we downsample the active parts of the feature map by a factor of 5, and for every incoming event, $\boldsymbol{u_i}$ the new events will have a new spatial location at $\boldsymbol{u_{i_{d}}}=(x_i/5,y_i/5)$. In the next processing step, we apply event-based max-pooling to extract the neighbouring $p\times p$ matrix ($p=3$) around the current event $\boldsymbol{u_{i_{d}}}$ from all the feature maps, resulting in a matrix of size $p\times p \times N$ for every incoming event. To construct the $X$ feature matrix, the pooled matrix is flattened to a 1D vector, resulting in $X \in \mathbb{R}^{E\times (p\times p \times N)}$ where $Y \in \mathbb{R}^{E\times 1}$. A cross-validation of $50 \ by \ 50$ is applied to split the features and labels matrices to train and test sets, $X_{train}$, $Y_{train}$, $X_{test}$ and $Y_{test_{gt}}$. To calculate the linear output weight $W_c$ that map between $X_{train}$, $Y_{train}$, the Online Pseudo-inverse Update Method (OPIUM)~\cite{van_schaik_online_2015} is used, as follows, 

\begin{equation}
   W_c=\frac{A^{-1}Y}{A^{-1}A+\alpha}\label{eq:opium}
\end{equation}

Where $A=X_{train}$, $Y=Y_{train}$ and $\alpha=1e-6$ is a regularization factor. $W_c$ is then used to predict the $Y_{test_{pred}}=W_c \cdot X_{test}$. Finally, we convert the continuous values in $Y_{test_{pred}}$ to binary using one-hot encoding to evaluate it with the binary $Y_{test_{gt}}$ which was modified where label=1 is satellite and label=0 is everything else (stars events were discarded).

\section{Experimental Results}

\subsection{Event-based Noise Filtering Baselines}

To evaluate the performance of our proposed algorithms: CrossConv, FEAST, and FEAST+classifier, we benchmarked them against 11 publicly available noise filtering algorithms. Below, we provide an overview of each of these baseline methods.

\textbf{KNoise~\cite{khodamoradi_on-space_2018}}. performs noise filtering by checking whether an event has a temporal correlation with its neighbouring events within a defined time window. It maintains two memory buffers to store the most recent events for each row and column in the sensor array.

\textbf{Fixed Window Filter (FWF)~\cite{guo_low_2023}}. assesses the spatiotemporal correlation of events by storing the spatial locations of the $L$ most recent events within a fixed-size window. For each new event, it computes the Manhattan distances to these previous events and identifies the minimum distance $D_{\text{min}}$. If $D_{\text{min}}$ exceeds a predefined threshold, the event is considered noise and discarded.

\textbf{Dual Window Filter (DWF)~\cite{guo_low_2023}}. improves upon FWF by utilizing two separate memory windows: one for signal events and another for noise events. It applies the same distance-based filtering rules as FWF within each window, enhancing the discrimination between signal and noise.

\textbf{Spatiotemporal Correlation Filter (STCF)~\cite{guo_low_2023}}. filters out events that lack support from the $k$ most recent events within a $3 \times 3$ spatial neighbourhood and within a temporal window defined by a time constant $\tau$.

\textbf{TS (Time Surface) Filter~\cite{lagorce_hots_2017}}. converts the event stream into a time surface representation using an exponentially decaying kernel. For noise filtering, it compares the timestamp of the current event with those of neighboring events within a certain radius. Events are retained if the temporal difference is below a pre-defined threshold, ensuring that only temporally coherent events are passed through.

\textbf{Motion-based Flow (EvFlow)~\cite{wang_ev-gait_2019}}. estimates optical flow by performing local plane fitting to compute gradients in the spatiotemporal event data. It then filters out events with abnormal flow values, based on the assumption that noise events will not conform to the consistent motion patterns of the scene.

\textbf{YNoise~\cite{feng_event_2020}}. algorithm computes the density of each incoming event in its spatiotemporal domain. Events located in high-density regions are considered part of moving edge and are retained, while those in low-density regions are filtered out as noise.

\textbf{Inceptive Events (IETS)~\cite{karray_inceptive_2019}}. classifies events into inceptive events and scaling events. Inceptive events are triggered by edges or corners in the scene and are detected based on their temporal isolation from preceding events. Scaling events, which represent the magnitude of the intensity change but provide less structural information, are discarded. The algorithm focuses on retaining inceptive events and considers scaling events as noise.

\textbf{Multilayer Perceptron Filter (MLPF)~\cite{guo_low_2023,Navaro2023-mlpf-dvs-denoising}}. is a lightweight, hardware-friendly neural network model with a single hidden layer. It learns spatiotemporal structural patterns from the data to discriminate between noise and signal events. The algorithm operates by extracting a patch of the nearest neighbouring pixels around each new event. These patches are processed using a linear decay kernel to capture temporal information, and the resulting features are fed into the classifier.

\textbf{AEDAT~\cite{fang_aednet_2022}}. is a deep learning model based on the PointNet architecture~\cite{qi2017pointnet}. It applies one-dimensional convolutions to neighbouring pixels along the time dimension, capturing the spatiotemporal connectivity of events while preserving their asynchronous nature.

\subsection{Experimental Details}

To process the stream of events for the logical-based algorithms, we ran each algorithm on all the recordings and calculated the average value in the evaluation. For the learning-based algorithms, we applied a cross-validation of 50 by 50 and used both event polarities. We train the models on 50\% of the dataset and performed the noise filtering during inference and evaluation on the other 50\%. To train the FEAST noise filtering algorithm, we randomised the order of the training set and ran through them over 10 epochs with 200 neurons for the background (i.e. noise and hot pixels) and 200 neurons for the foreground (i.e. stars and satellites) with a $11 \times 11$ event context. The time constant was set to $\tau = 1 \times 10^5 \mu s$, with positive and negative thresholds of $\theta^{+} = 0.001$ and $\theta^{-} = 0.002$, respectively. To account for the speed differences between the foreground and background, we selected different learning rates: $\eta = 0.0005$ for the background and $\eta = 0.01$ for the foreground. The same trained weights were used in the FEAST+classifier architecture.

For the MLPF algorithm, we first construct a \(17 \times 17\) event patch centred around each event, storing the timestamp and polarity of each neighbouring event. This results in a \(17 \times 17 \times 2\) array for each event. For each recording, we generate a \texttt{.csv} file for every second of data, containing the timestamp and polarity information of the patches. During training, we set the learning rate to \(\eta = 0.0001\) and used a batch size of 100 with 500 hidden neurons. The time constant was set to \(\tau = 1000\,\mu\mathrm{s}\), and we applied a dropout rate of 0.2. The model was trained for 100 epochs. We used the same default parameters for the AEDAT algorithm~\footnote{https://github.com/Fanghuachen/AEDNet} for training and inference.

All algorithms are implemented in Python. We ran all algorithms on an RTX 3080 with 64 cores at 3.5GHz, on an Ubuntu 20.04 LTS system.

\textbf{Evaluation Metrics}. We formulated the noise-filtering task as a binary classification problem between signal and noise, where the signal represents satellite events and the noise comprises background noise and hot pixels. During training, all events were utilized, but the performance evaluation of the algorithms focused exclusively on satellites, hot pixels, and background noise. To fairly compare the algorithms, we used the ROC curve~\cite{guo_low_2023,Navaro2023-mlpf-dvs-denoising} as an accuracy measure, to show the True Positive Rate (TPR) and False Positive Rate (FPR) over all thresholds, providing a clear picture of the effect on signal and noise discrimination. This is also to prevent the bias of showing a single accuracy value when dealing with multiple algorithms. An ideal noise filtering algorithm achieves FPR=0 which indicates that all noise events are removed and TPR=1 which indicates that all satellite events are preserved. To further evaluate the performance of all algorithms, we looked at the ROC curve and picked the parameter that gave the lowest FPR and the highest TPR to measure Signal Retain (SR), Noise Removal (NR), Hot Pixels Removal (HPR) and Denoising Accuracy (DA)~\cite{duan_led_2024} as well as the Area Under Curve (AUC). The metrics are described as follows:

\begin{equation}
TPR = \frac{\sum_{i=1}^{N} \left( d_i = 1 \land g_i = 1 \right)}{\sum_{i=1}^{N} \left( g_i = 1 \right)} \\
\end{equation}

\begin{equation}
FPR = \frac{\sum_{i=1}^{N} \left( d_i = 1 \land g_i = 0 \right)}{\sum_{i=1}^{N} \left( g_i = 0 \right)}
\end{equation}

\begin{equation}
SR = 100 \times \frac{\sum_{i=1}^{N} \left( d_i = 1 \land g_i = 1 \right)}{\sum_{i=1}^{N} \left( g_i = 1 \right)} \\
\end{equation}

\begin{equation}
NR = 100 \times \frac{\sum_{i=1}^{N} \left( d_i = 0 \land g_i = 0 \right)}{\sum_{i=1}^{N} \left( g_i = 0 \right)} \\
\end{equation}

\begin{equation}
HPR = 100 \times \frac{\sum_{i=1}^{N} \left( d_i = 0 \land g_i = 2 \right)}{\sum_{i=1}^{N} \left( g_i = 2 \right)} \\
\end{equation}

\begin{equation}
DA = \frac{SR+NR+HPR}{3}
\end{equation}

where $\land$ denotes the logical AND operation, $N$ total number of events, $d_i$ is the noise filtering output with 1s represent satellites events and 0s represent noise+hot pixels events, and $g_i$ is ground truth with 0s represent the noise events, 1s represents satellite events and 2s represents hot pixels.

The SR metric assesses the effectiveness of preserving satellite events (i.e. same as TPR), the NR metric evaluates the removal of background noise, and the HPR metric measures evaluate the removal of hot pixel events. DA combines these aspects by averaging SR, NR, and HPR, offering a single metric that reflects the algorithm's overall ability to retain useful signals while removing unwanted noise.

To generate the ROC curve for the logic-based algorithms, we performed a parameter sweep over the threshold, following the approach in~\cite{guo_low_2023}. For KNoise~\cite{khodamoradi_on-space_2018}, TS~\cite{lagorce_hots_2017}, FWF~\cite{guo_low_2023}, and DWF~\cite{guo_low_2023}, we varied the distance threshold parameter while keeping other settings constant. Specifically, for FWF and DWF, we used a window length of $L=175$. For TS, we set the decay time to $1e6$ and used a search radius of 5. For KNoise, the duration was fixed at 1 second. For YNoise~\cite{feng_event_2020}, EvFlow~\cite{wang_ev-gait_2019}, and STDF~\cite{feng_event_2020}, we swept the duration parameter. The search radius was set to 5 for YNoise and 1 for EvFlow, while for STDF, we used a correlation number of 3. For STCF~\cite{guo_low_2023}, we swept the correlation time parameter while fixing the correlation number to 4. For FEAST we swept the time decay. For IETS~\cite{karray_inceptive_2019} we swept the windowsize. For MLPF~\cite{guo_low_2023}, AEDAT~\cite{fang_aednet_2022} and FEAST+classifier, we swept the classification threshold from 0 to 1.

\subsection{Experimental Results}

\textbf{Quantitative Evaluations}. Table~\ref{tab:quanteval} presents the quantitative results averaged over all test sets, comparing various noise filtering methods on the Ev-Satellite dataset. Among the logic-based algorithms, EvFlow~\cite{wang_ev-gait_2019} achieves the highest signal retain (i.e. SR) rate and second highest AUC. However, its performance in hot pixel removal (i.e. HPR) is comparatively lower. In contrast, CrossConv excels in removing hot pixels, achieving the best result in this category among logic-based methods, while also maintaining a high SR rate. However, its noise removal (i.e. NR) rate is lower.

\begin{table}[h]
\centering
\caption{Quantitative results comparison on the Ev-Satellite dataset on all logical and learning-based event-based noise filtering algorithms. We mark the \textbf{best} in bold and \underline{second best}.}
\label{tab:quanteval}
\small %
\setlength{\tabcolsep}{4pt} %
\begin{tabular}{lccccc}
\Xhline{3\arrayrulewidth}
Metrics & AUC$\uparrow$ & SR$\uparrow$ & NR$\uparrow$ & HPR$\uparrow$ & DA$\uparrow$ \\ \hline
\multicolumn{6}{c}{\textcolor{red}{\textbf{Logical-based Algorithms}}} \\ %
KNoise~\cite{khodamoradi_on-space_2018} & 0.77 & 55.05 & 98.99 & \underline{98.72} & 84.25 \\
FWF~\cite{guo_low_2023} &  0.66 & 62.89 & 59.23 & 51.95 & 58.03 \\
DWF~\cite{guo_low_2023} & 0.67 & 66.78 & 53.27 & 47.70 & 55.92 \\
STDF~\cite{feng_event_2020} & 0.48 & 81.08 & 34.34 & 34.00 & 49.80 \\
TS~\cite{lagorce_hots_2017} & 0.83 & 76.15 & 88.52 & 89.14 & 84.60 \\
EvFlow~\cite{wang_ev-gait_2019}  & \underline{0.98} & \textbf{95.98} & \underline{99.73} & 58.72 & 84.81 \\
YNoise~\cite{feng_event_2020} & 0.69 & 92.08 & 81.17 & 81.26 & 84.84 \\
IETS~\cite{karray_inceptive_2019} & 0.02 & 20.85 & 84.78 & 84.78 & 63.47 \\
STCF~\cite{guo_low_2023} & 0.42 & 26.74 & 90.24 & 86.78 & 67.92 \\
\rowcolor{lightgray} %
\textbf{CrossConv} & 0.80 & \underline{95.93} & 34.92 & 90.29 & 73.71 \\ \hline
\multicolumn{6}{c}{\textcolor{red}{\textbf{Learning-based Algorithms}}} \\ %
MLPF~\cite{guo_low_2023} & 0.83 & 88.87 & 86.92 & 89.08 & 88.29 \\
AEDNet~\cite{fang_aednet_2022} & \underline{0.98} & 67.59 & 37.82 & 35.63 & 47.02 \\
\rowcolor{lightgray} %
\textbf{FEAST}~\cite{afshar_event-based_2020} & \textbf{0.99} & 85.54 & 99.00 & 98.29 & \textbf{94.28} \\
\rowcolor{lightgray} %
\textbf{FEAST + Classifier}~\cite{afshar_event-based_2020} & \underline{0.98} & 82.49 & \textbf{99.84} & \textbf{99.88} & \underline{94.07} \\
\Xhline{3\arrayrulewidth}
\end{tabular}
\end{table}

\begin{figure}[h]
\begin{center}
   \includegraphics[width=\linewidth]{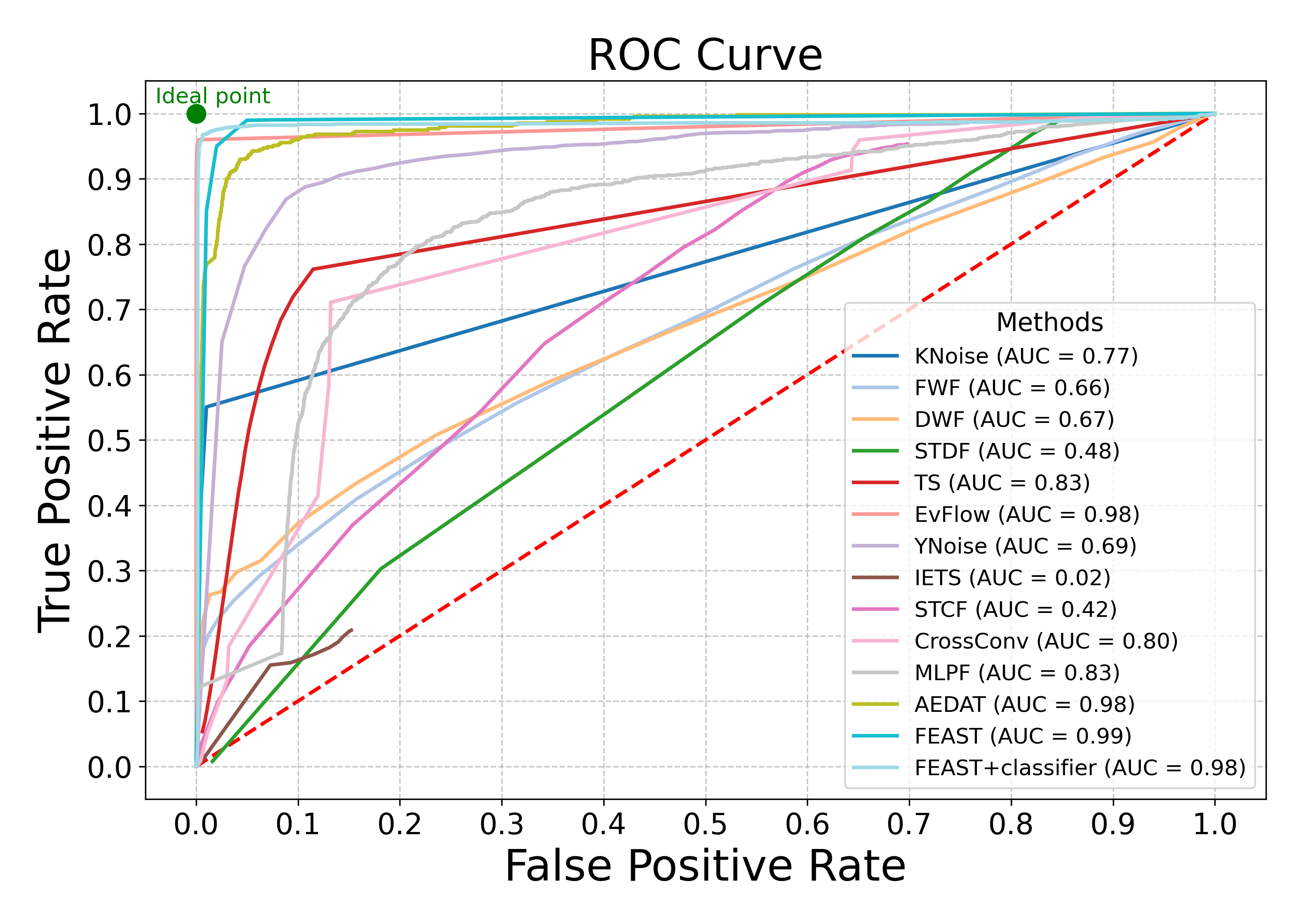}\\[1em]
\end{center}
   \caption{ROC curves and summarized AUC values on the Ev-Satellites datasets.}
\label{fig:roccurve}
\end{figure}

Among the learning-based methods, the combination of FEAST with a classifier achieves the highest performance in NR and HPR, ranking second only to FEAST in overall denoising accuracy. The FEAST algorithm attains a significantly higher ROC/AUC than any of the noise filtering methods as shown in Figure~\ref{fig:roccurve}, suggesting it can learn more discriminative features to distinguish satellites from noise and hot pixels, leading to higher classification accuracy.

These results suggest that FEAST-based algorithms can learn highly discriminative features to distinguish satellites from noise and hot pixels, leading to higher classification accuracy. The higher AUC scores of learning-based methods indicate their superior ability to preserve very sparse satellite events while effectively removing unwanted noise.

\setlength{\fboxrule}{0.1pt} %
\begin{figure*}[htbp]\hspace*{-0.7cm}
\renewcommand*{\arraystretch}{0.1}\centering
\begin{subfigure}{\textwidth}
\begin{tabular}{c c c c c c c c c}
    & \small\color{gray!90}Input 1 & \small\color{gray!90}KNoise~\cite{khodamoradi_on-space_2018} & \small\color{gray!90}FWF~\cite{guo_low_2023} & \small\color{gray!90}DWF~\cite{guo_low_2023} & \small\color{gray!90}STDF~\cite{feng_event_2020} & \small\color{gray!90}TS~\cite{lagorce_hots_2017} & \small\color{gray!90}EvFlow~\cite{wang_ev-gait_2019} & \small\color{gray!90}YNoise~\cite{feng_event_2020} \\
    \tiny\rotatebox{90}{\hspace{.2cm}\color{gray!90}Denoised} & 
    \fbox{\includegraphics[height=0.45in,width=0.6in]{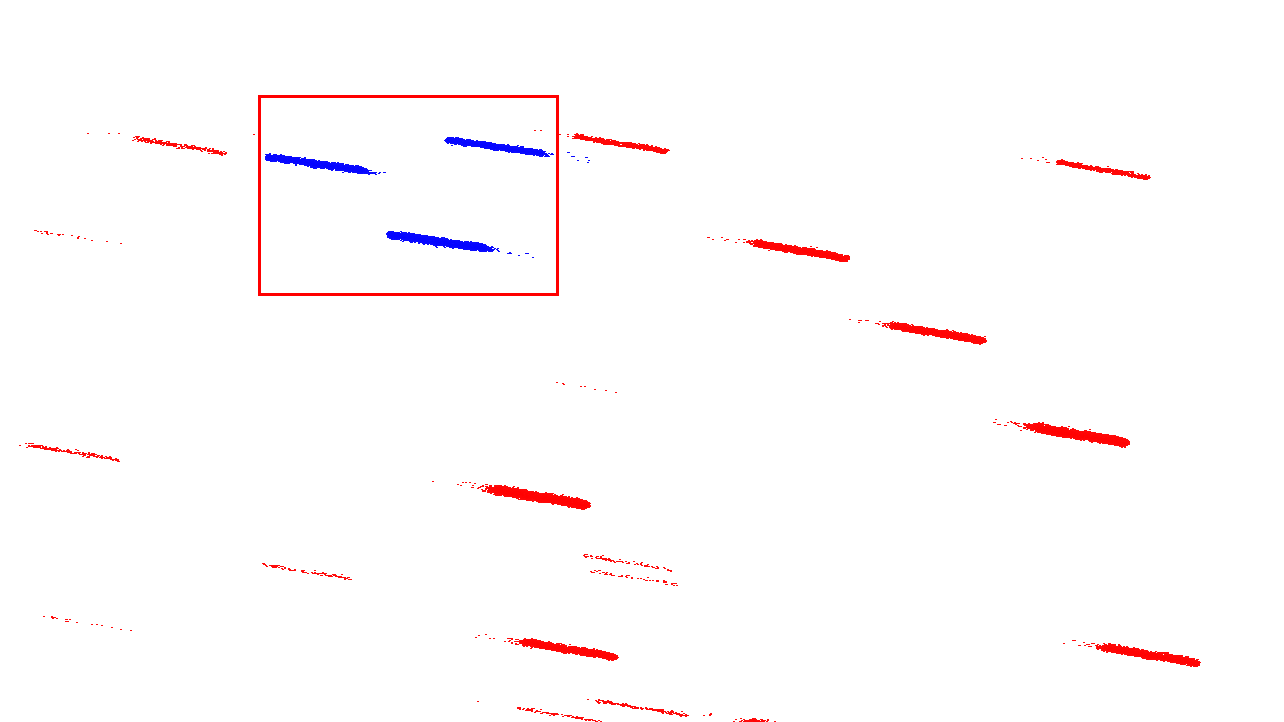}} & 
    \fbox{\includegraphics[height=0.45in,width=0.6in]{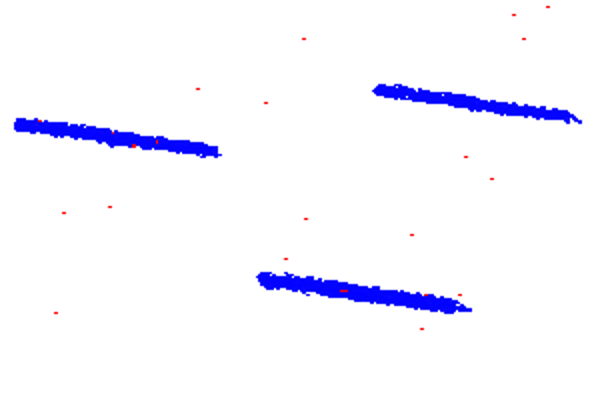}} &
    \fbox{\includegraphics[height=0.45in,width=0.6in]{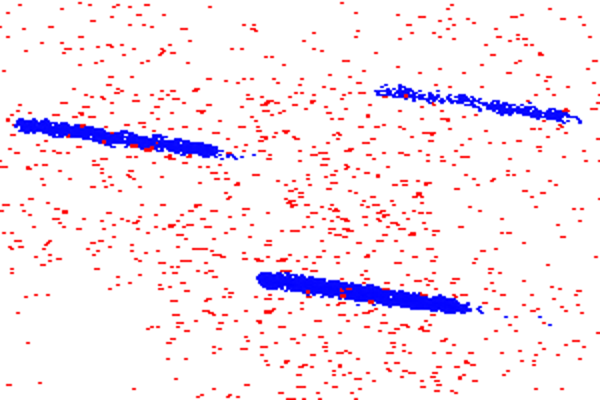}} &
    \fbox{\includegraphics[height=0.45in,width=0.6in]{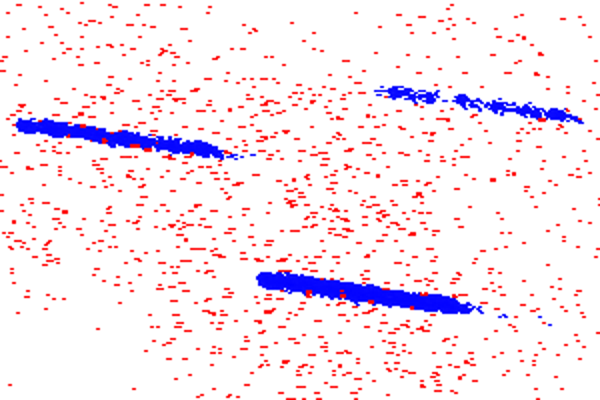}} &
    \fbox{\includegraphics[height=0.45in,width=0.6in]{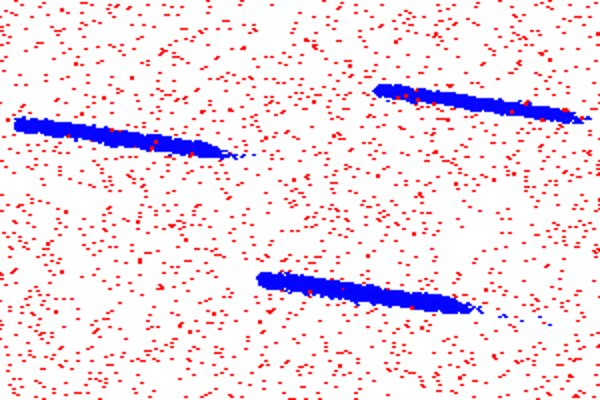}} &
    \fbox{\includegraphics[height=0.45in,width=0.6in]{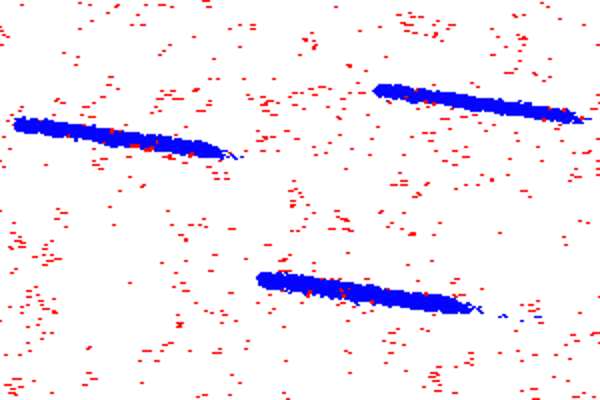}} &
    \fbox{\includegraphics[height=0.45in,width=0.6in]{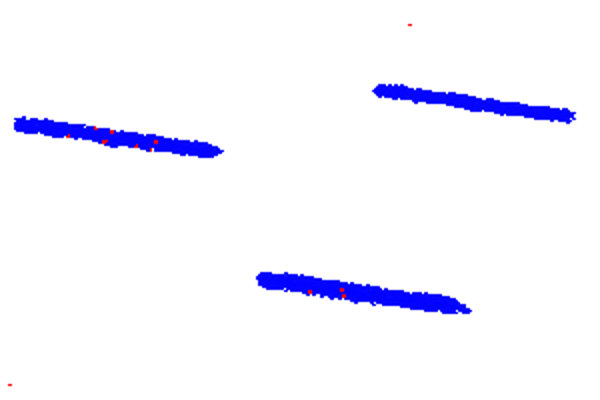}} &
    \fbox{\includegraphics[height=0.45in,width=0.6in]{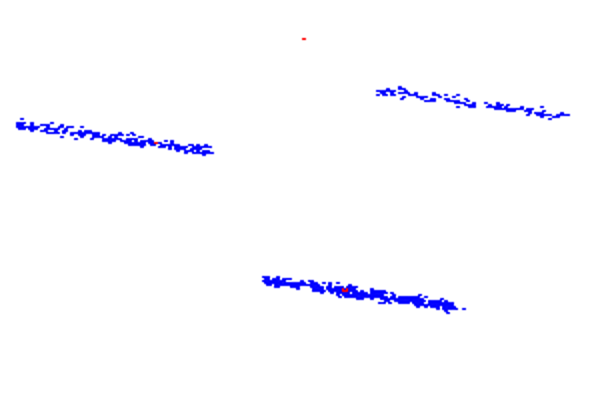}} \\
    \tiny\rotatebox{90}{\color{gray!90}Residual Noise} &
    \fbox{\includegraphics[height=0.45in,width=0.6in]{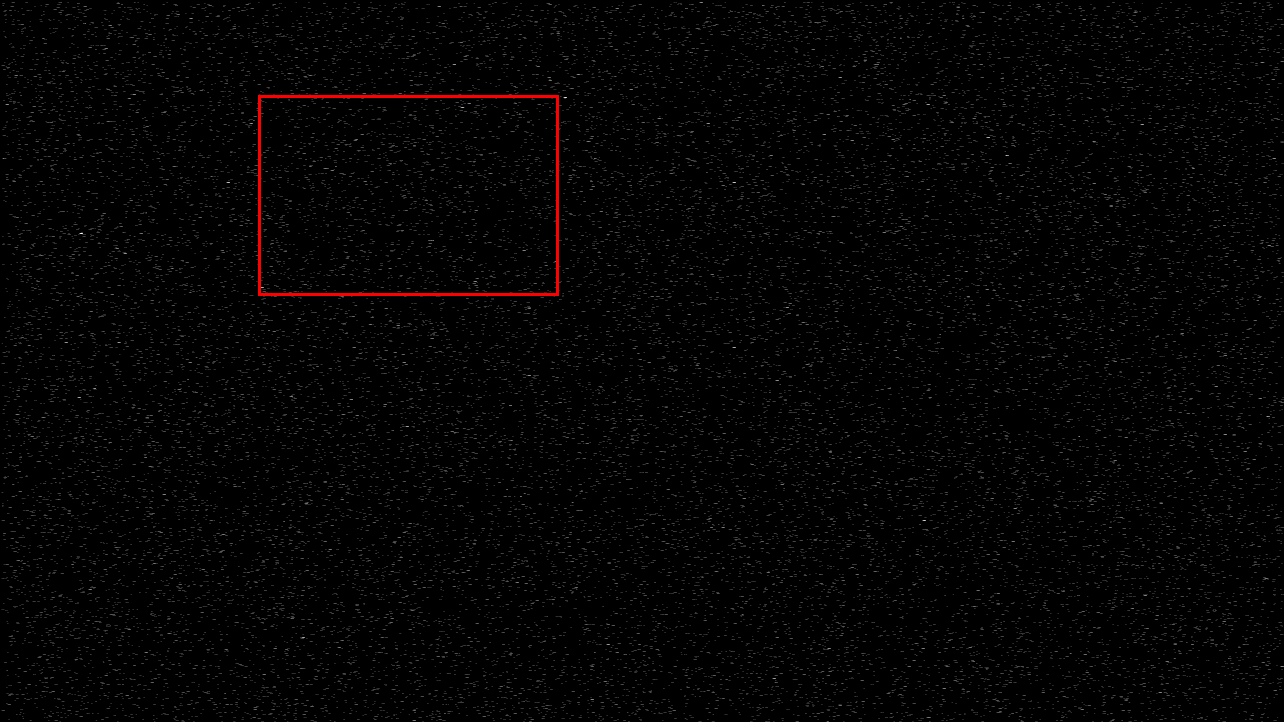}} & 
    \fbox{\includegraphics[height=0.45in,width=0.6in]{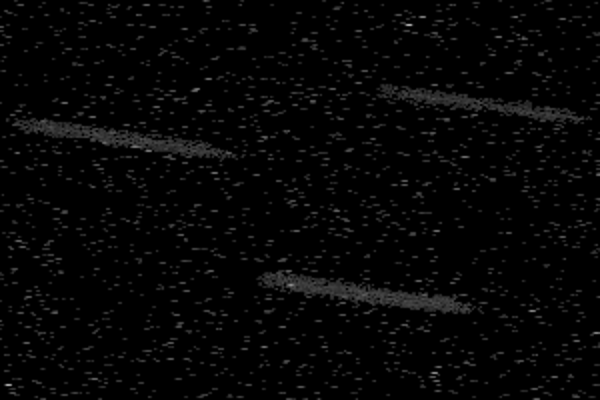}} &
    \fbox{\includegraphics[height=0.45in,width=0.6in]{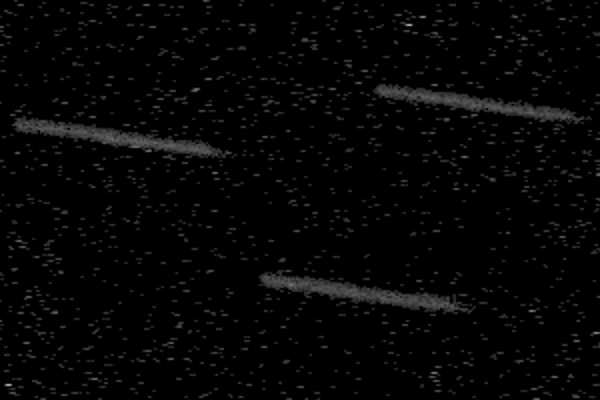}} &
    \fbox{\includegraphics[height=0.45in,width=0.6in]{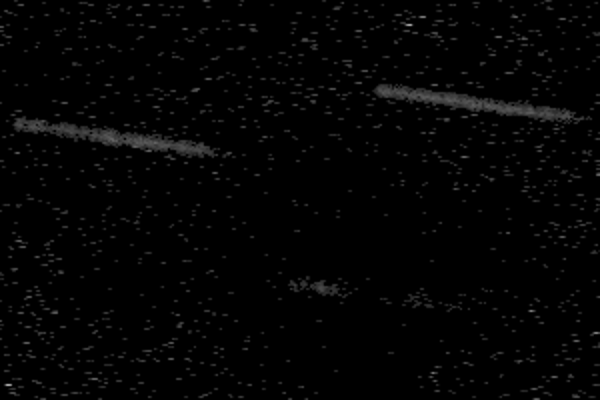}} &
    \fbox{\includegraphics[height=0.45in,width=0.6in]{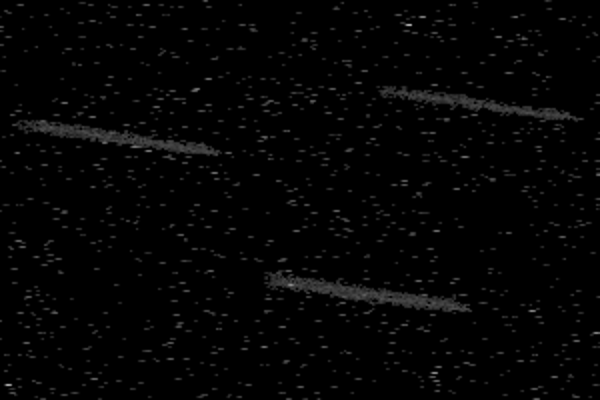}} &
    \fbox{\includegraphics[height=0.45in,width=0.6in]{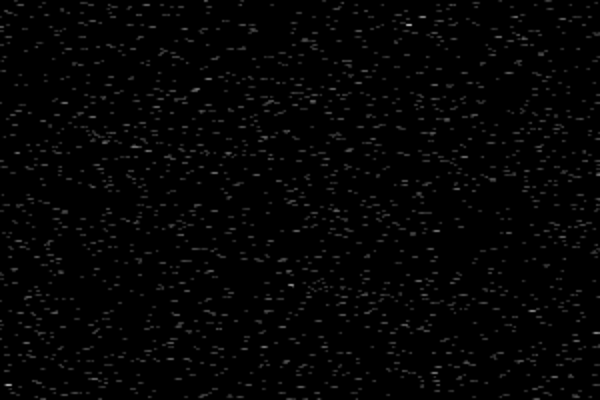}} &
    \fbox{\includegraphics[height=0.45in,width=0.6in]{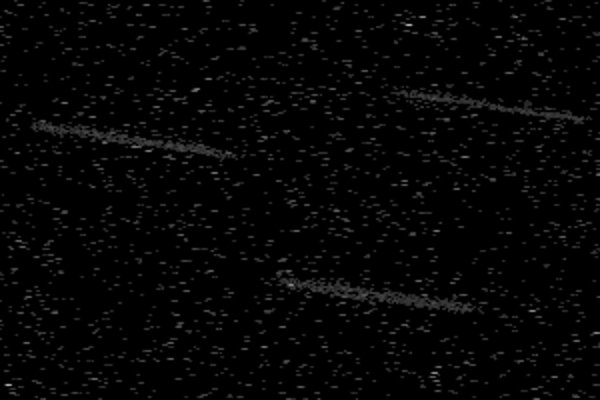}} &
    \fbox{\includegraphics[height=0.45in,width=0.6in]{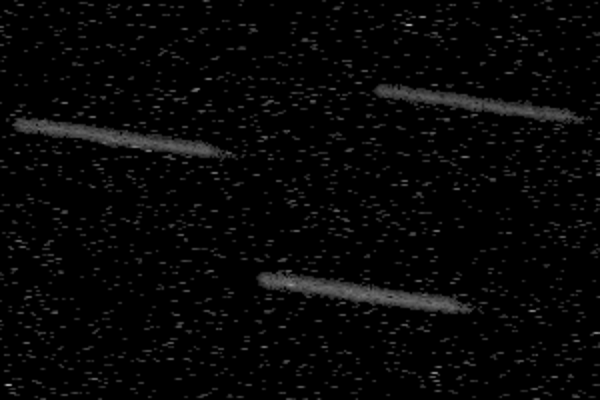}} \\
    & \small\color{gray!90}IETS~\cite{karray_inceptive_2019} & \small\color{gray!90}STCF~\cite{guo_low_2023} & \small\color{gray!90}CrossConv & \small\color{gray!90}MLPF~\cite{guo_low_2023} & \small\color{gray!90}AEDNet~\cite{fang_aednet_2022} & \small\color{gray!90}FEAST~\cite{afshar_event-based_2020} & \small\color{gray!90}FEAST+C~\cite{afshar_event-based_2020}\\
    \tiny\rotatebox{90}{\hspace{.2cm}\color{gray!90}Denoised} & 
    \fbox{\includegraphics[height=0.45in,width=0.6in]{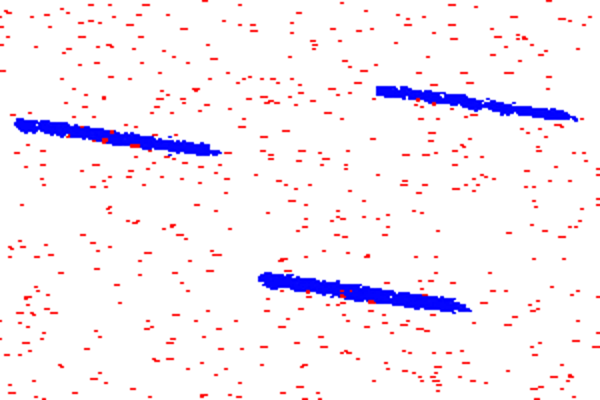}} & 
    \fbox{\includegraphics[height=0.45in,width=0.6in]{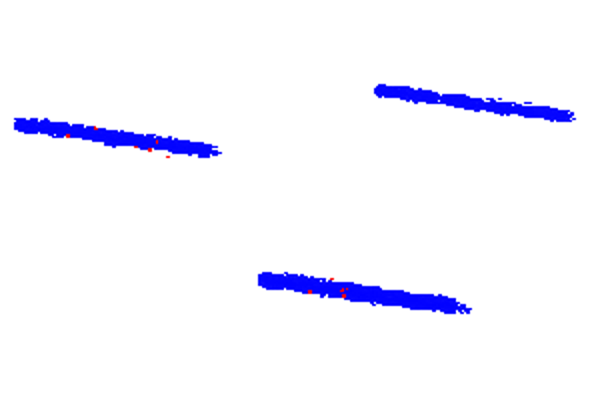}} & 
    \fbox{\includegraphics[height=0.45in,width=0.6in]{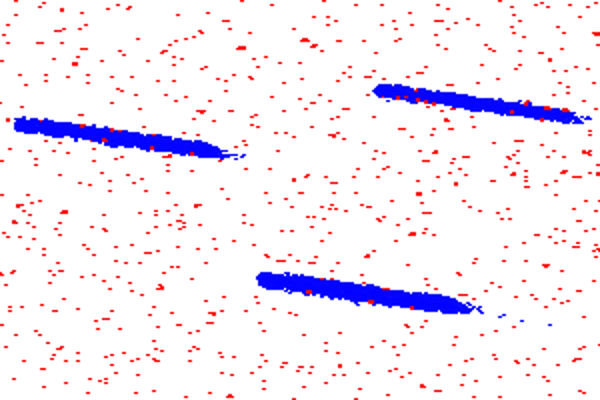}} & 
    \fbox{\includegraphics[height=0.45in,width=0.6in]{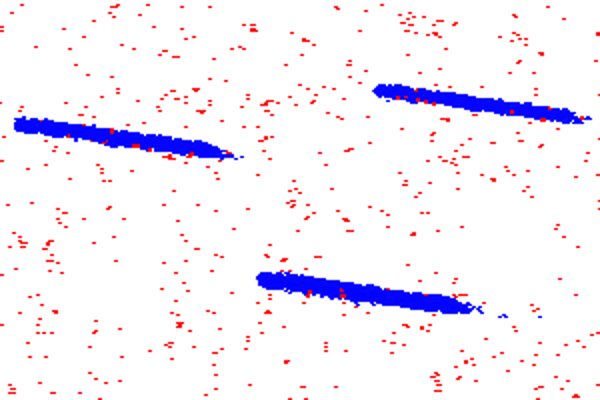}} & 
    \fbox{\includegraphics[height=0.45in,width=0.6in]{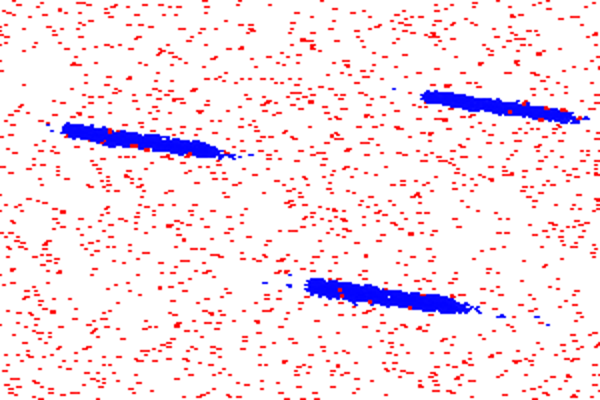}} & 
    \fbox{\includegraphics[height=0.45in,width=0.6in]{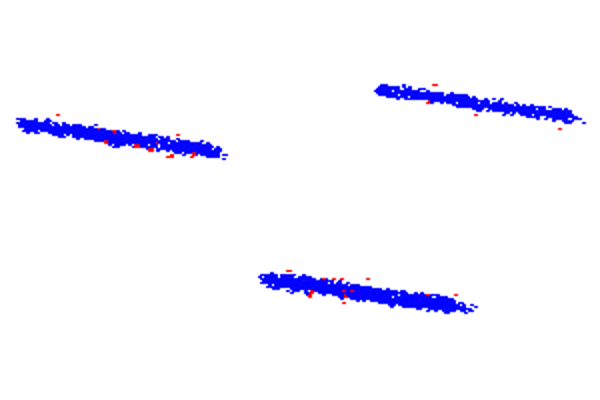}} & 
    \fbox{\includegraphics[height=0.45in,width=0.6in]{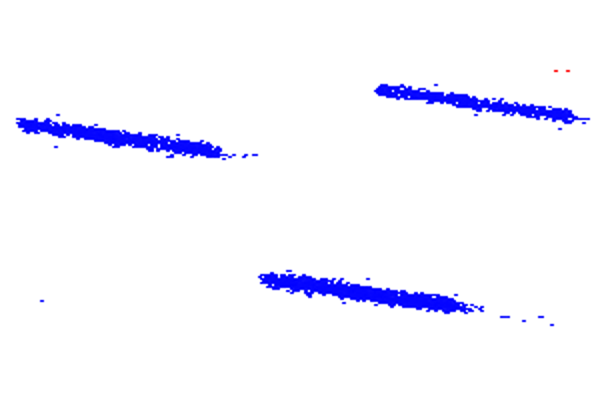}} \\
    \tiny\rotatebox{90}{\color{gray!90}Residual Noise} &
    \fbox{\includegraphics[height=0.45in,width=0.6in]{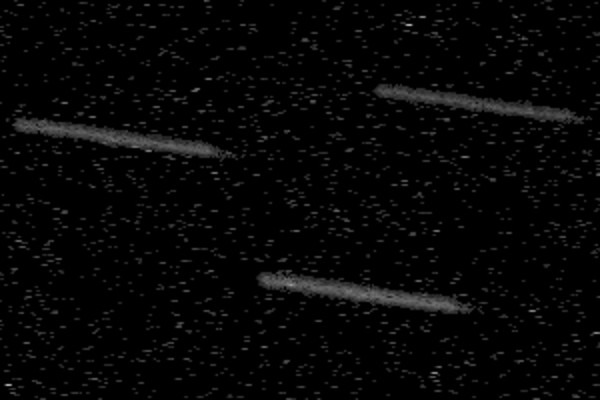}} & 
    \fbox{\includegraphics[height=0.45in,width=0.6in]{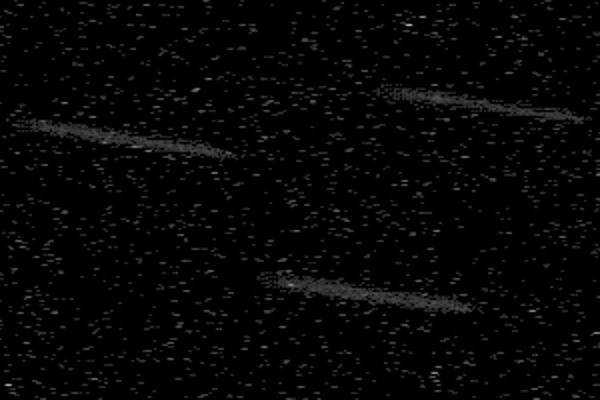}} & 
    \fbox{\includegraphics[height=0.45in,width=0.6in]{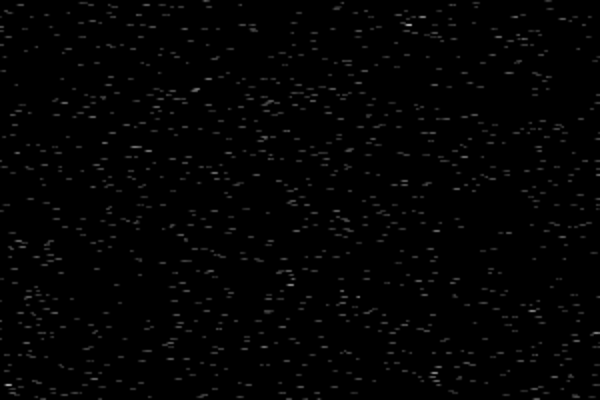}} & 
    \fbox{\includegraphics[height=0.45in,width=0.6in]{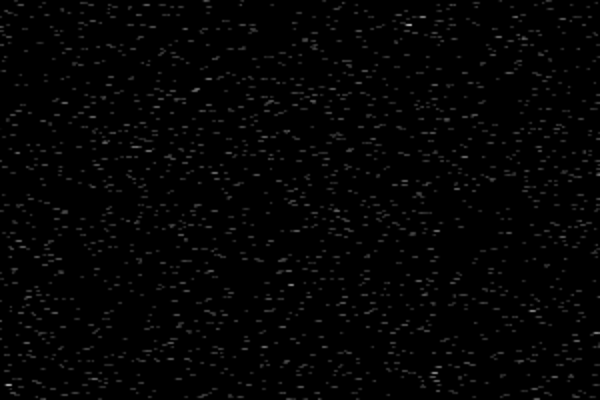}} & 
    \fbox{\includegraphics[height=0.45in,width=0.6in]{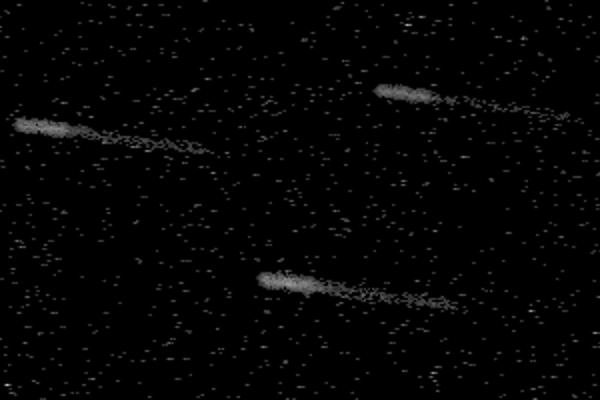}} & 
    \fbox{\includegraphics[height=0.45in,width=0.6in]{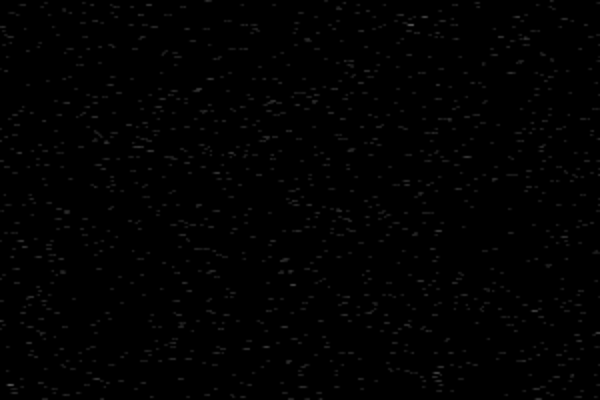}} & 
    \fbox{\includegraphics[height=0.45in,width=0.6in]{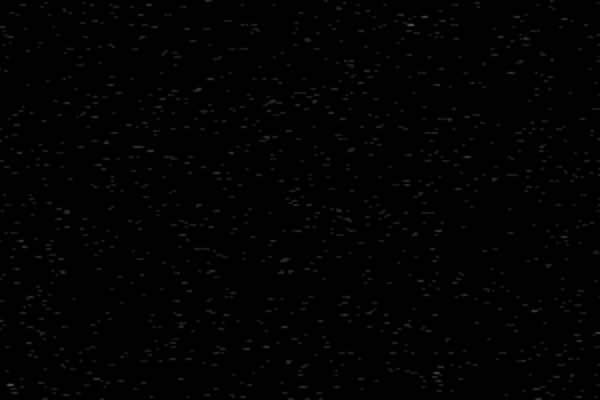}} \\
\end{tabular}
\end{subfigure}
\hspace*{-0.7cm}
    \renewcommand*{\arraystretch}{0.7}
    \begin{subfigure}{\textwidth}
        \centering
        \begin{tabular}{c c c c c c c c c}
    & \small\color{gray!90}Input 2 & \small\color{gray!90}KNoise~\cite{khodamoradi_on-space_2018} & \small\color{gray!90}FWF~\cite{guo_low_2023} & \small\color{gray!90}DWF~\cite{guo_low_2023} & \small\color{gray!90}STDF~\cite{feng_event_2020} & \small\color{gray!90}TS~\cite{lagorce_hots_2017} & \small\color{gray!90}EvFlow~\cite{wang_ev-gait_2019} & \small\color{gray!90}YNoise~\cite{feng_event_2020} \\
    \tiny\rotatebox{90}{\hspace{.2cm}\color{gray!90}Denoised} & 
    \fbox{\includegraphics[height=0.45in,width=0.6in]{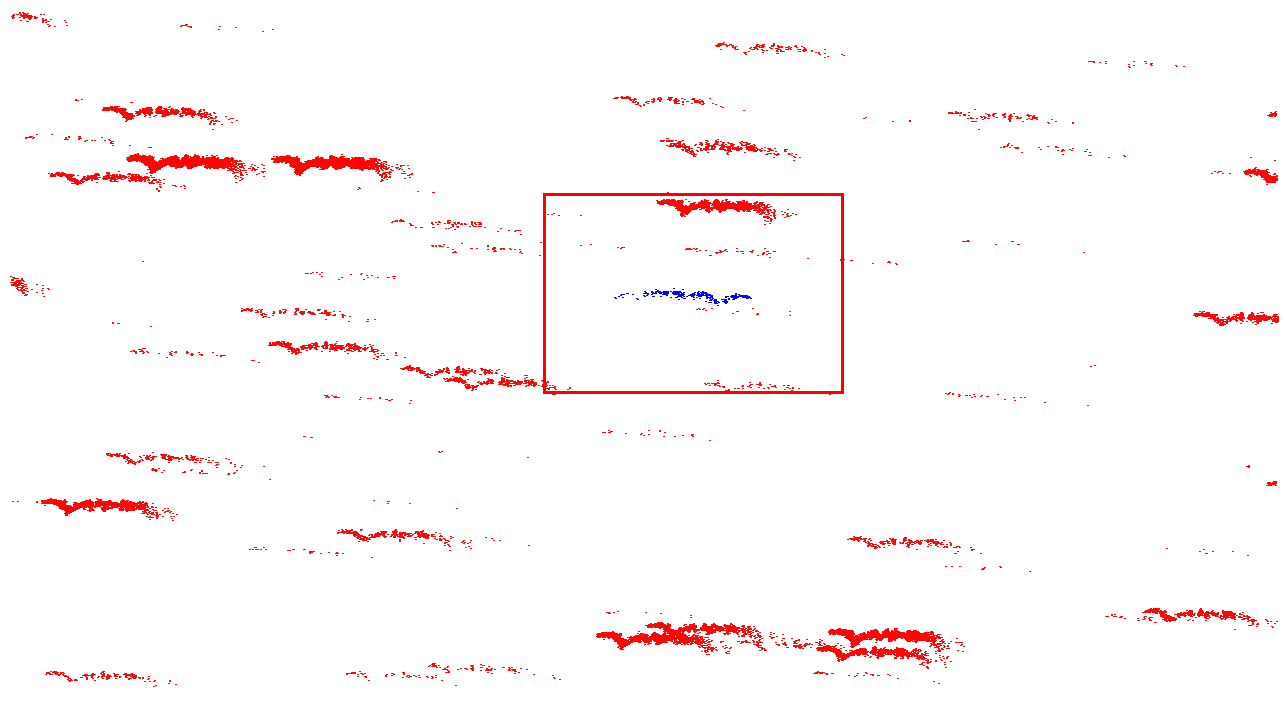}} & 
    \fbox{\includegraphics[height=0.45in,width=0.6in]{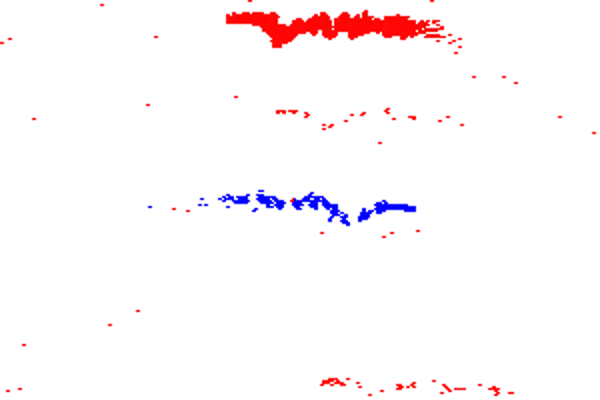}} &
    \fbox{\includegraphics[height=0.45in,width=0.6in]{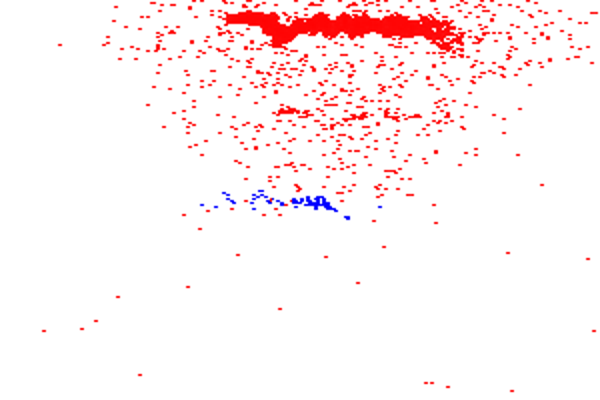}} &
    \fbox{\includegraphics[height=0.45in,width=0.6in]{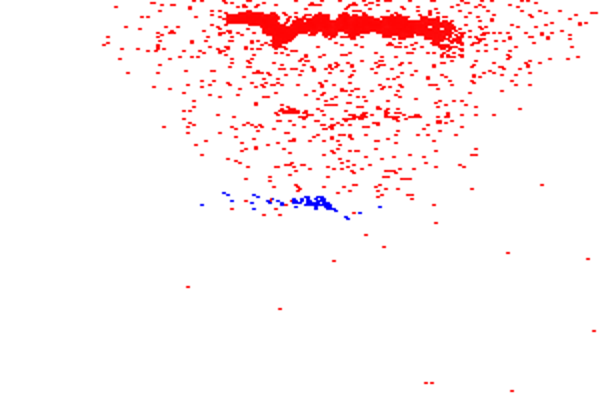}} &
    \fbox{\includegraphics[height=0.45in,width=0.6in]{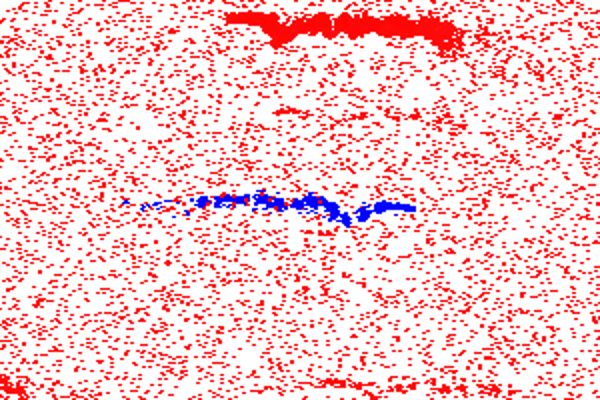}} &
    \fbox{\includegraphics[height=0.45in,width=0.6in]{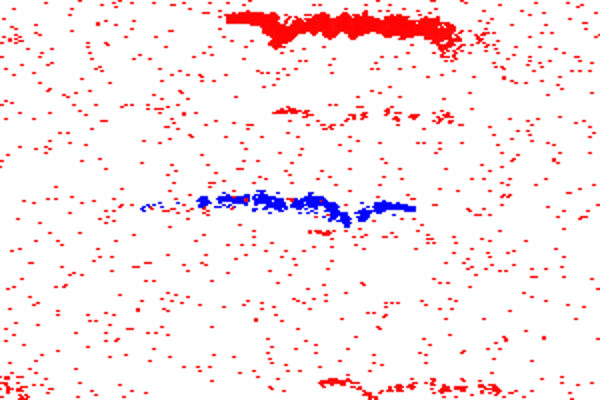}} &
    \fbox{\includegraphics[height=0.45in,width=0.6in]{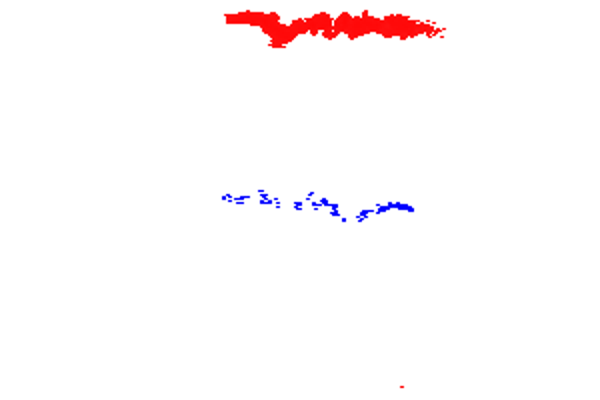}} &
    \fbox{\includegraphics[height=0.45in,width=0.6in]{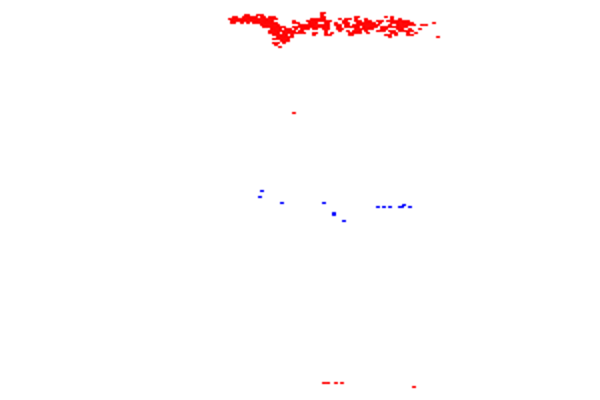}} \\
    \tiny\rotatebox{90}{\color{gray!90}Residual Noise} &
    \fbox{\includegraphics[height=0.45in,width=0.6in]{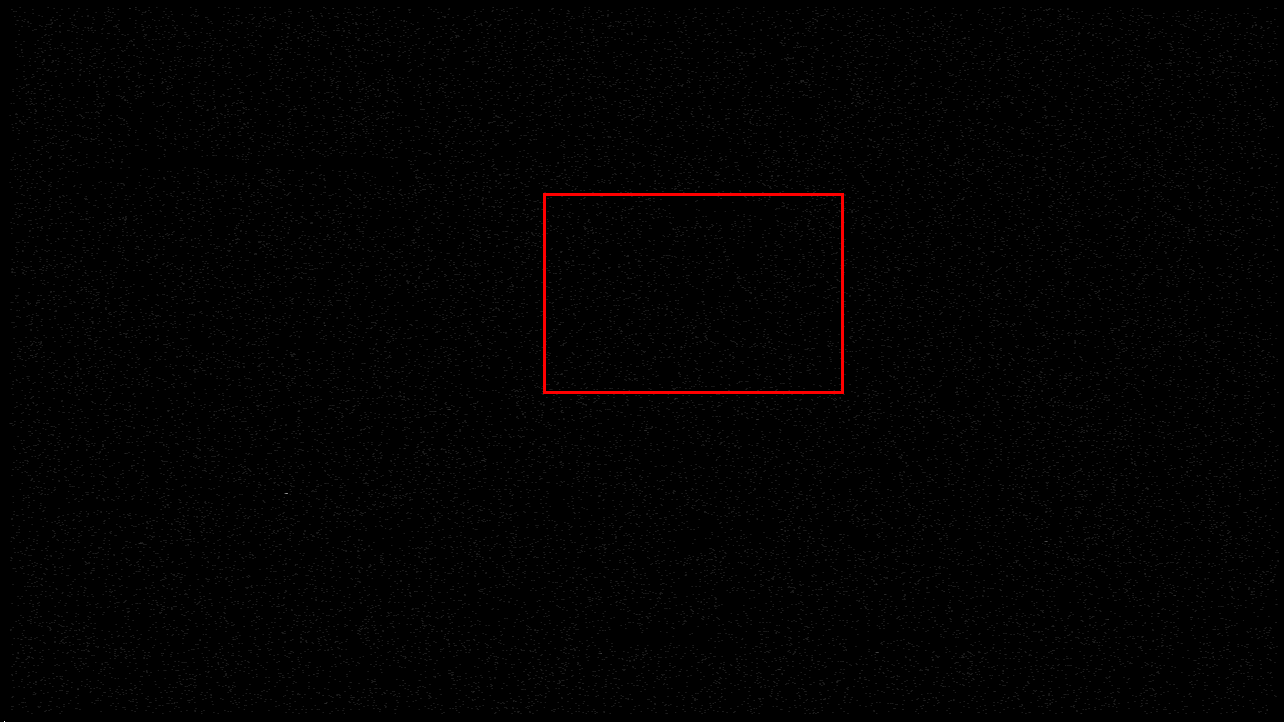}} & 
    \fbox{\includegraphics[height=0.45in,width=0.6in]{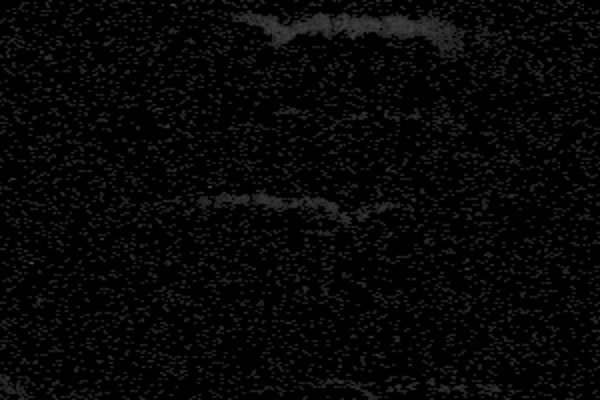}} &
    \fbox{\includegraphics[height=0.45in,width=0.6in]{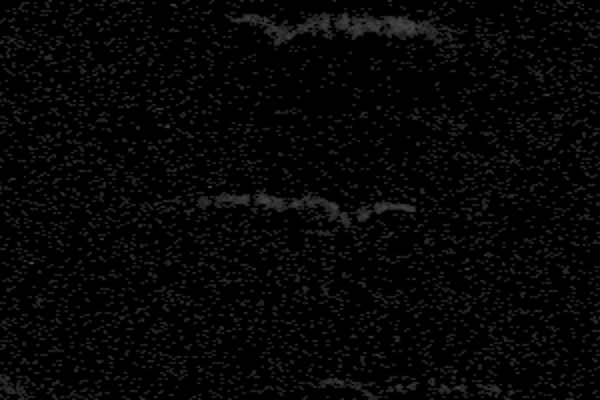}} &
    \fbox{\includegraphics[height=0.45in,width=0.6in]{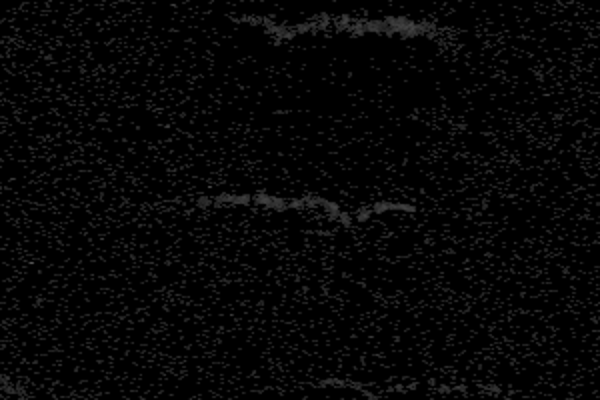}} &
    \fbox{\includegraphics[height=0.45in,width=0.6in]{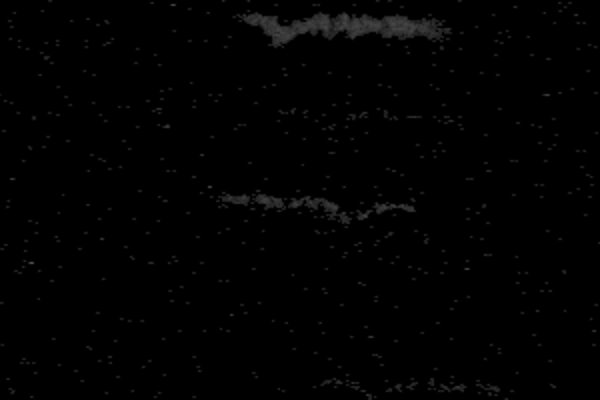}} &
    \fbox{\includegraphics[height=0.45in,width=0.6in]{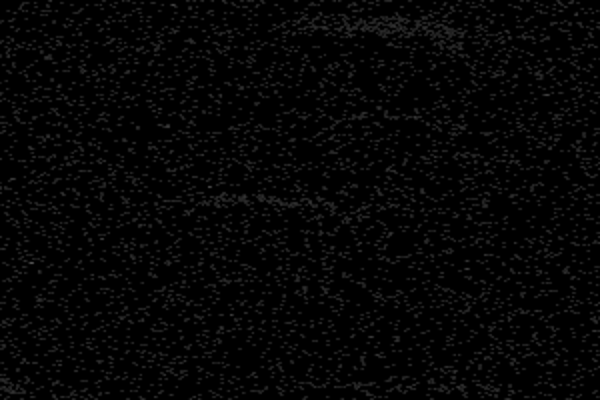}} &
    \fbox{\includegraphics[height=0.45in,width=0.6in]{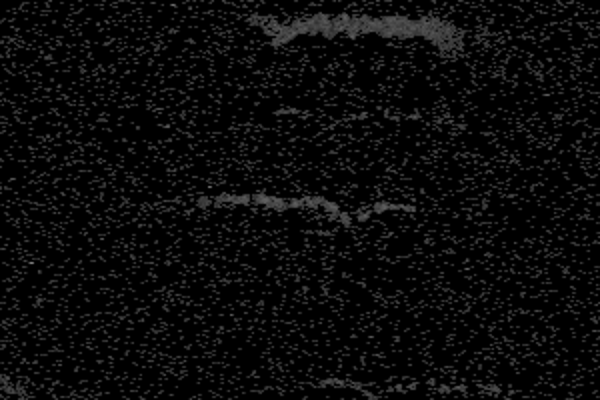}} &
    \fbox{\includegraphics[height=0.45in,width=0.6in]{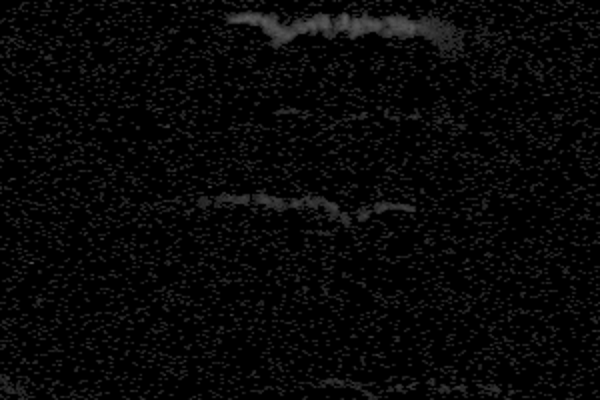}} \\
    & \small\color{gray!90}IETS~\cite{karray_inceptive_2019} & \small\color{gray!90}STCF~\cite{guo_low_2023} & \small\color{gray!90}CrossConv & \small\color{gray!90}MLPF~\cite{guo_low_2023} & \small\color{gray!90}AEDNet~\cite{fang_aednet_2022} & \small\color{gray!90}FEAST~\cite{afshar_event-based_2020} & \small\color{gray!90}FEAST+C~\cite{afshar_event-based_2020}\\
    \tiny\rotatebox{90}{\hspace{.2cm}\color{gray!90}Denoised} & 
    \fbox{\includegraphics[height=0.45in,width=0.6in]{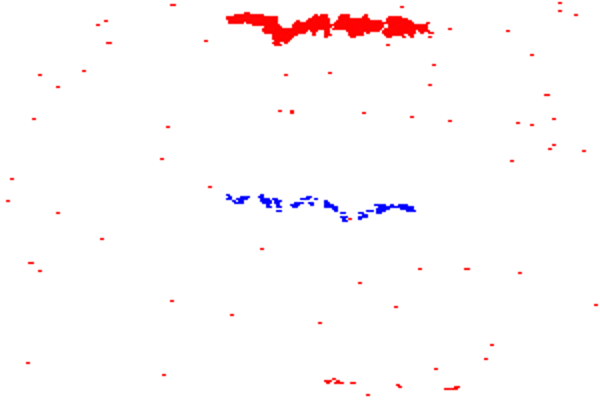}} & 
    \fbox{\includegraphics[height=0.45in,width=0.6in]{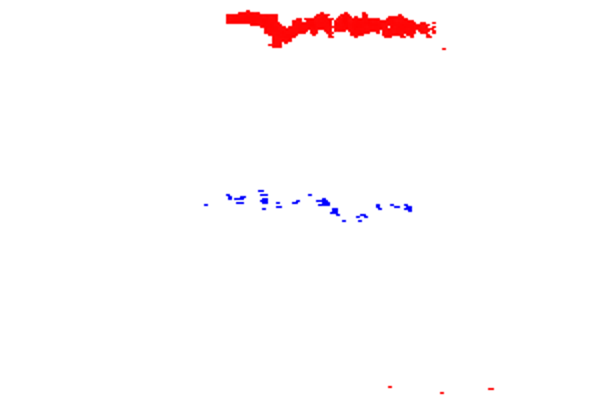}} & 
    \fbox{\includegraphics[height=0.45in,width=0.6in]{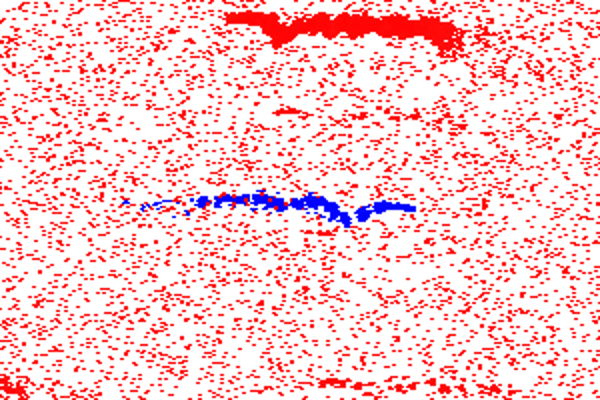}} & 
    \fbox{\includegraphics[height=0.45in,width=0.6in]{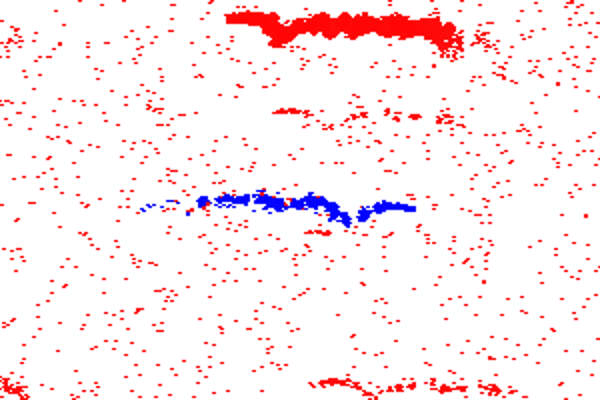}} & 
    \fbox{\includegraphics[height=0.45in,width=0.6in]{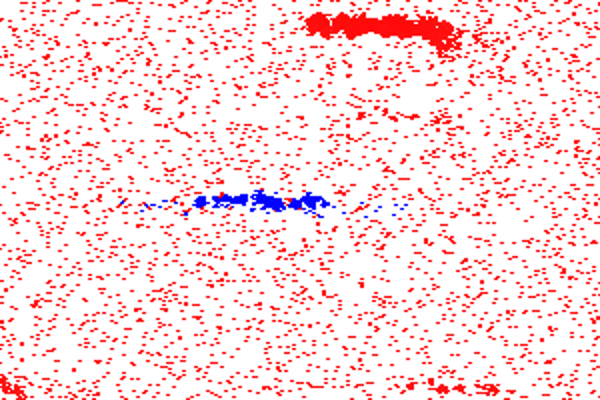}} & 
    \fbox{\includegraphics[height=0.45in,width=0.6in]{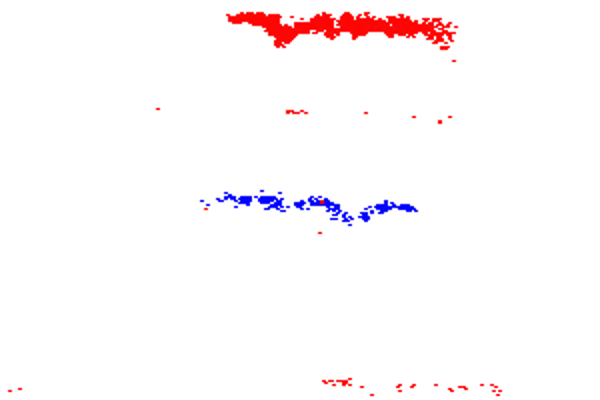}} & 
    \fbox{\includegraphics[height=0.45in,width=0.6in]{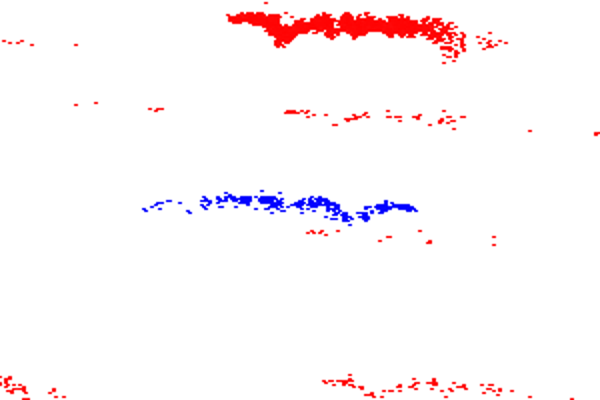}} \\
    \tiny\rotatebox{90}{\color{gray!90}Residual Noise} &
    \fbox{\includegraphics[height=0.45in,width=0.6in]{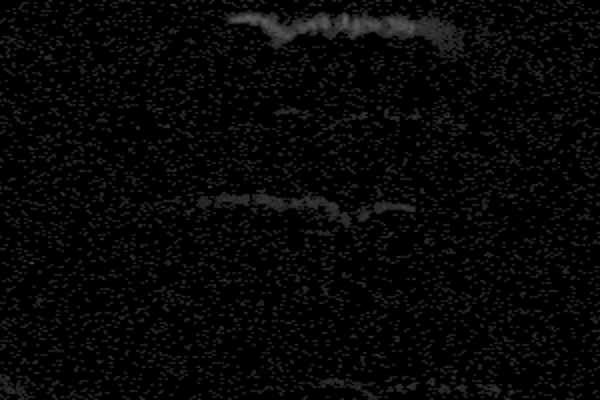}} & 
    \fbox{\includegraphics[height=0.45in,width=0.6in]{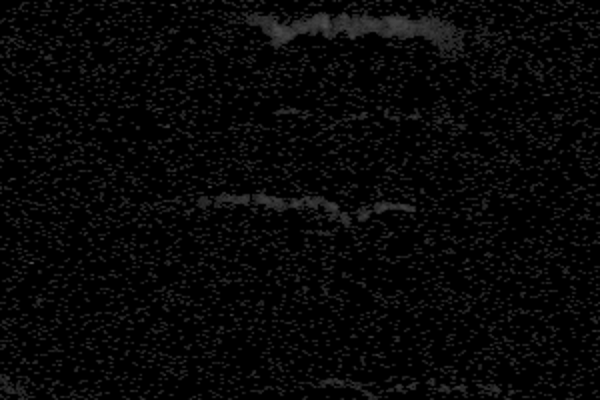}} & 
    \fbox{\includegraphics[height=0.45in,width=0.6in]{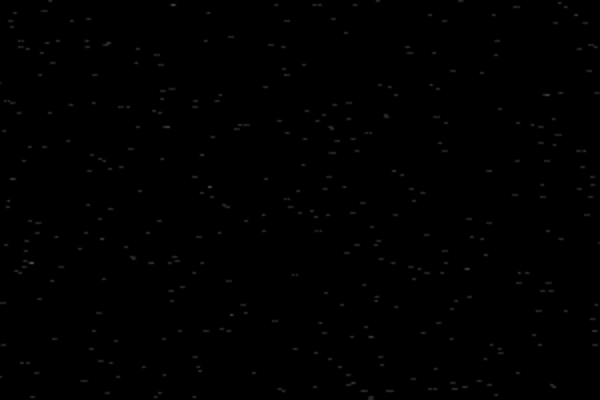}} & 
    \fbox{\includegraphics[height=0.45in,width=0.6in]{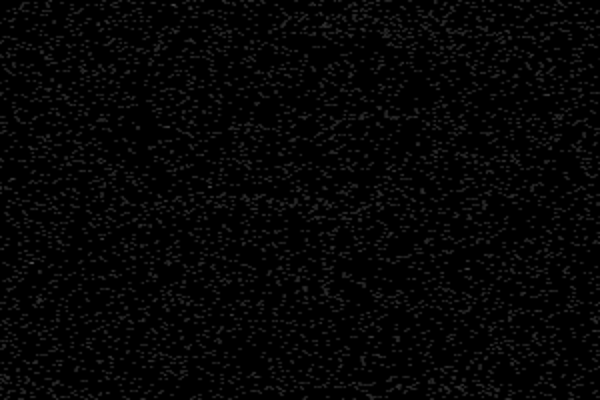}} & 
    \fbox{\includegraphics[height=0.45in,width=0.6in]{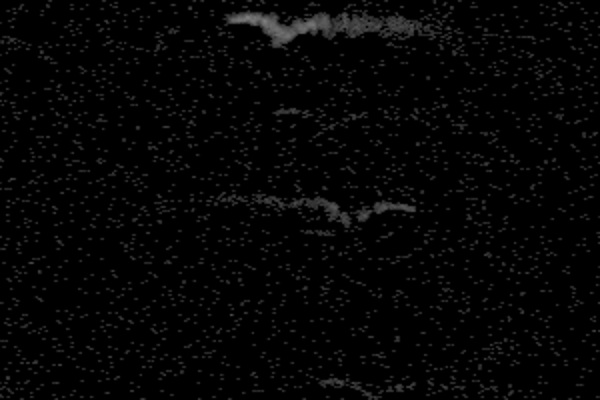}} & 
    \fbox{\includegraphics[height=0.45in,width=0.6in]{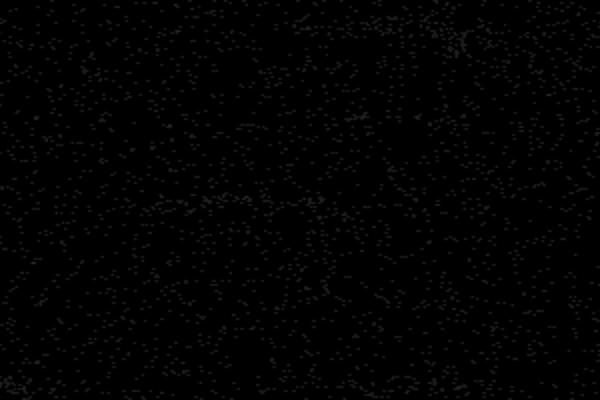}} & 
    \fbox{\includegraphics[height=0.45in,width=0.6in]{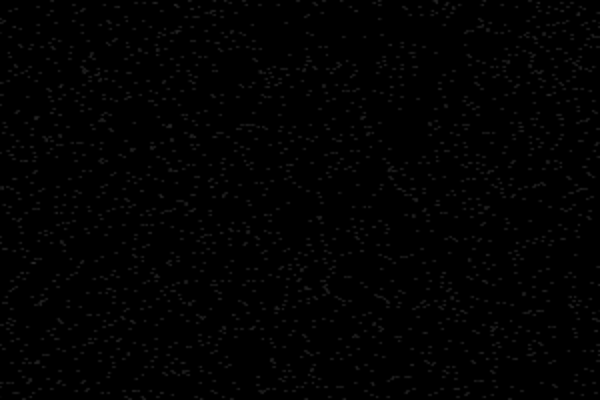}} \\
\end{tabular}
\end{subfigure}
\hspace*{-0.7cm}
    \renewcommand*{\arraystretch}{0.7}
    \begin{subfigure}{\textwidth}
        \centering
        \begin{tabular}{c c c c c c c c c}
    & \small\color{gray!90}Input 3 & \small\color{gray!90}KNoise~\cite{khodamoradi_on-space_2018} & \small\color{gray!90}FWF~\cite{guo_low_2023} & \small\color{gray!90}DWF~\cite{guo_low_2023} & \small\color{gray!90}STDF~\cite{feng_event_2020} & \small\color{gray!90}TS~\cite{lagorce_hots_2017} & \small\color{gray!90}EvFlow~\cite{wang_ev-gait_2019} & \small\color{gray!90}YNoise~\cite{feng_event_2020} \\
    \tiny\rotatebox{90}{\hspace{.2cm}\color{gray!90}Denoised} & 
    \fbox{\includegraphics[height=0.45in,width=0.6in]{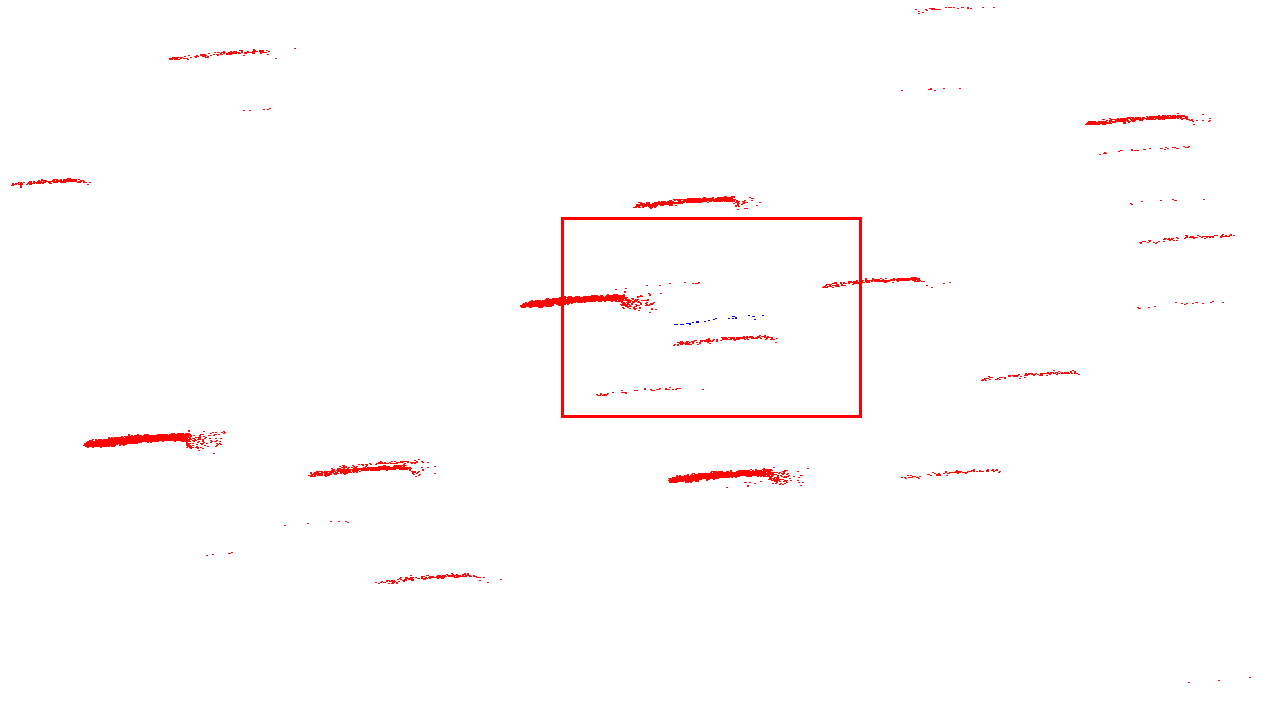}} & 
    \fbox{\includegraphics[height=0.45in,width=0.6in]{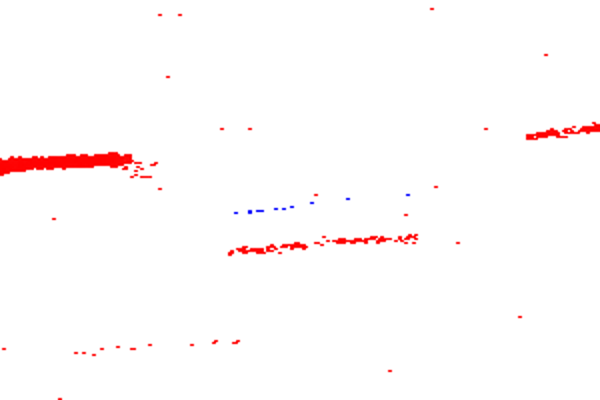}} &
    \fbox{\includegraphics[height=0.45in,width=0.6in]{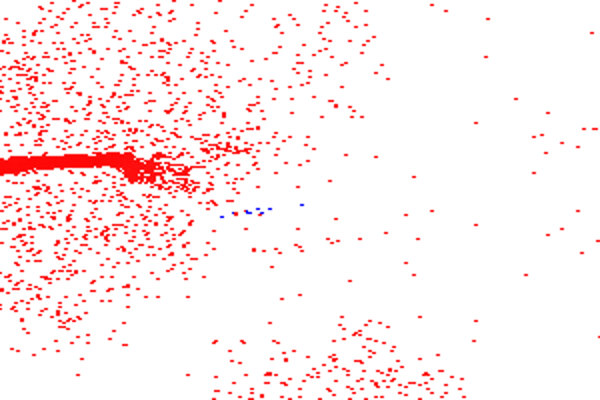}} &
    \fbox{\includegraphics[height=0.45in,width=0.6in]{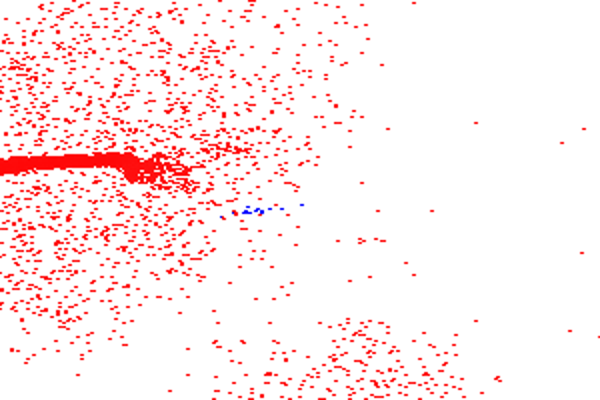}} &
    \fbox{\includegraphics[height=0.45in,width=0.6in]{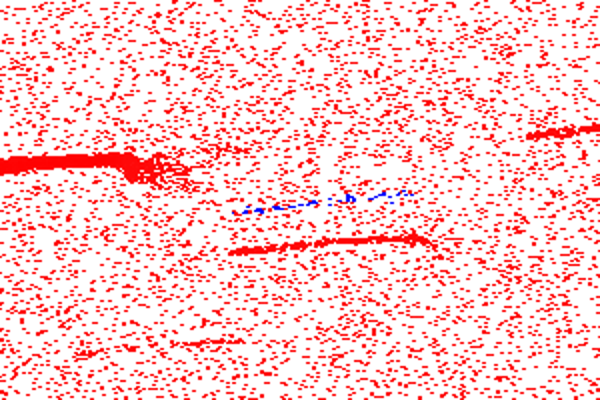}} &
    \fbox{\includegraphics[height=0.45in,width=0.6in]{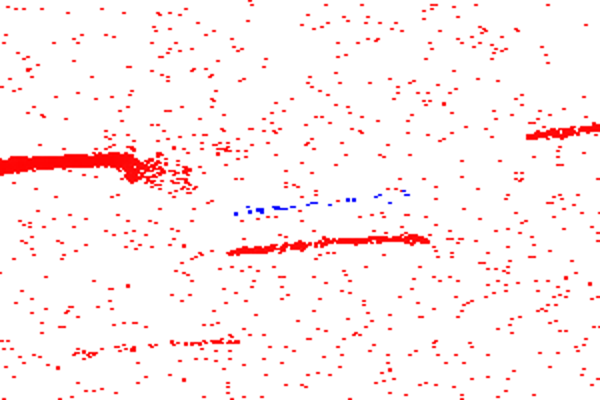}} &
    \fbox{\includegraphics[height=0.45in,width=0.6in]{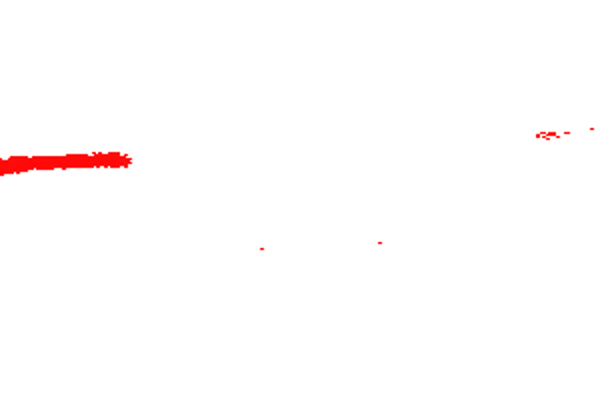}} &
    \fbox{\includegraphics[height=0.45in,width=0.6in]{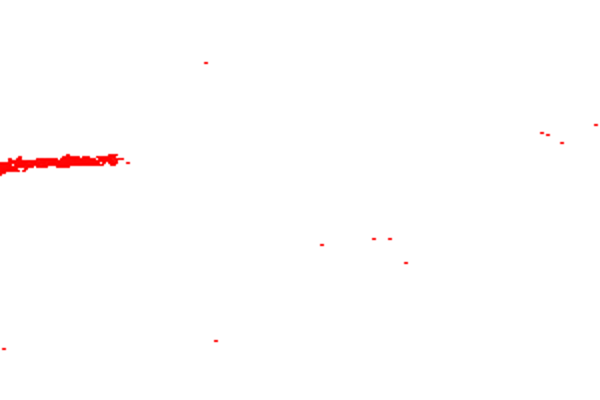}} \\
    \tiny\rotatebox{90}{\color{gray!90}Residual Noise} &
    \fbox{\includegraphics[height=0.45in,width=0.6in]{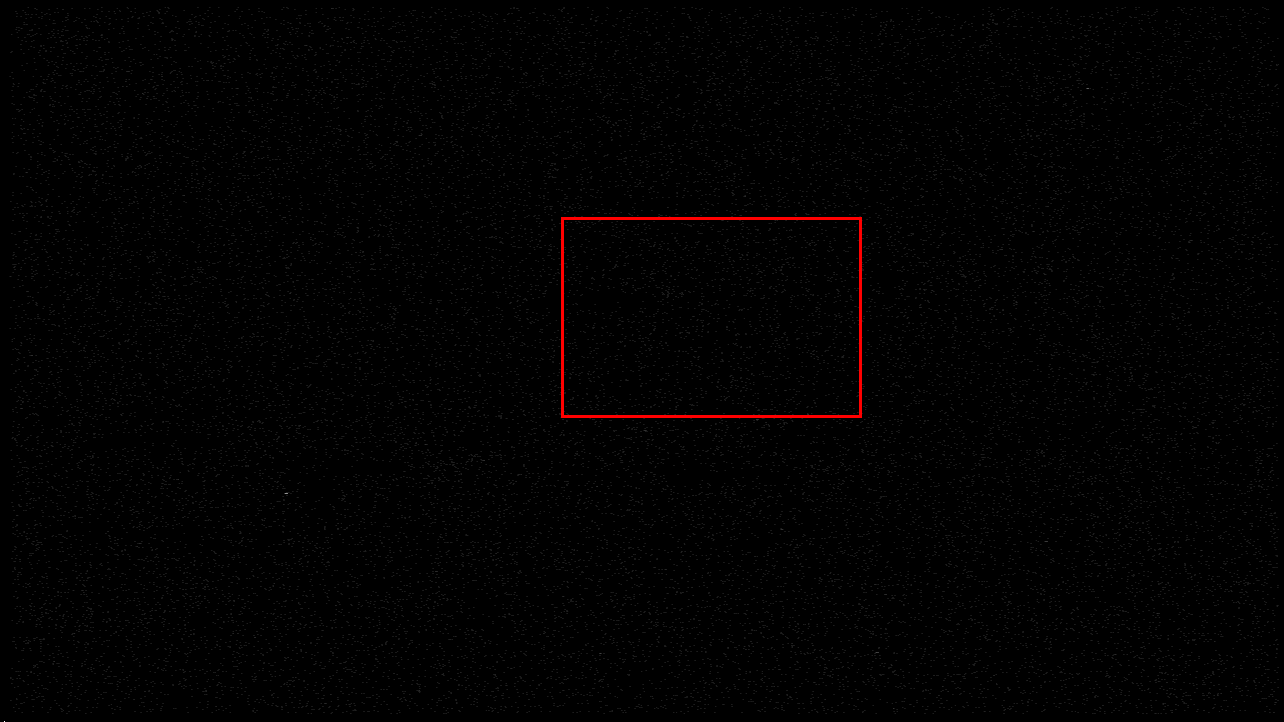}} & 
    \fbox{\includegraphics[height=0.45in,width=0.6in]{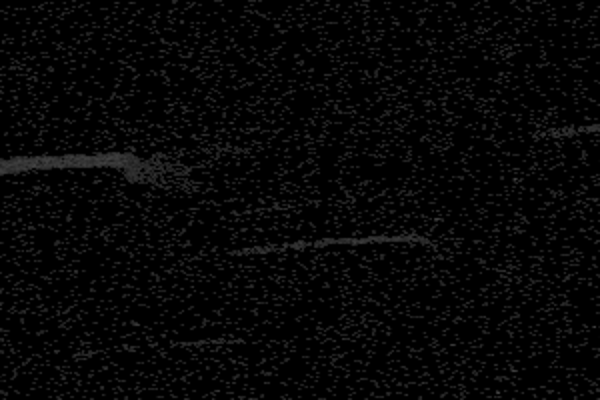}} &
    \fbox{\includegraphics[height=0.45in,width=0.6in]{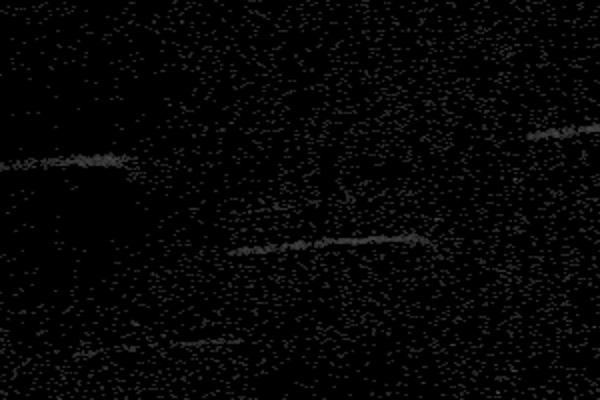}} &
    \fbox{\includegraphics[height=0.45in,width=0.6in]{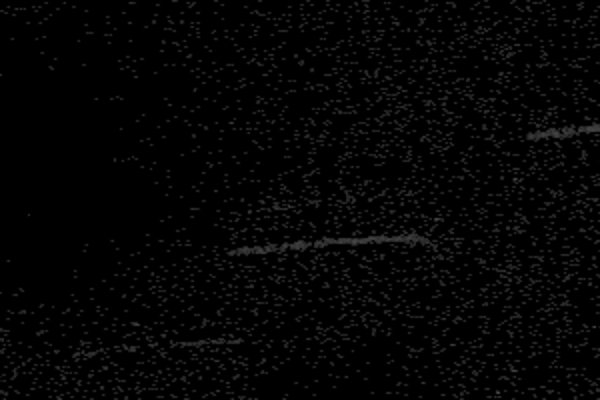}} &
    \fbox{\includegraphics[height=0.45in,width=0.6in]{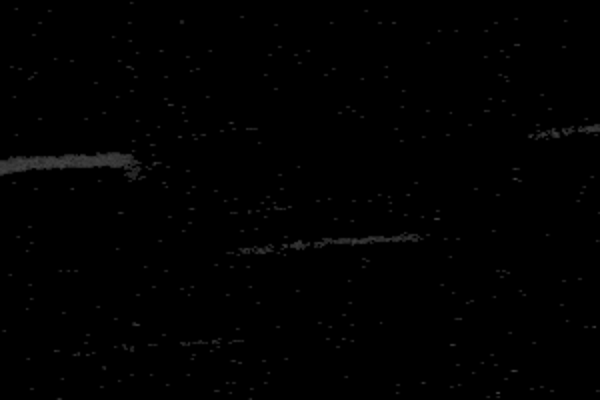}} &
    \fbox{\includegraphics[height=0.45in,width=0.6in]{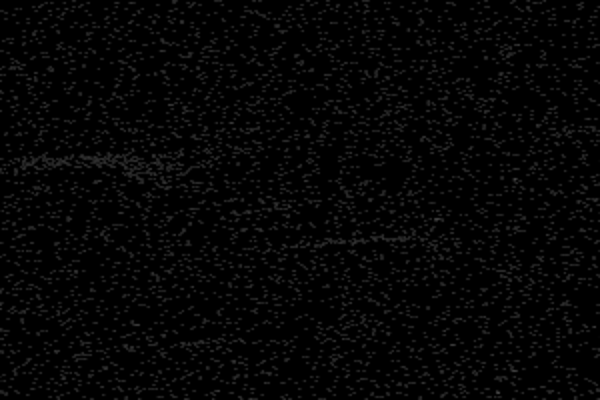}} &
    \fbox{\includegraphics[height=0.45in,width=0.6in]{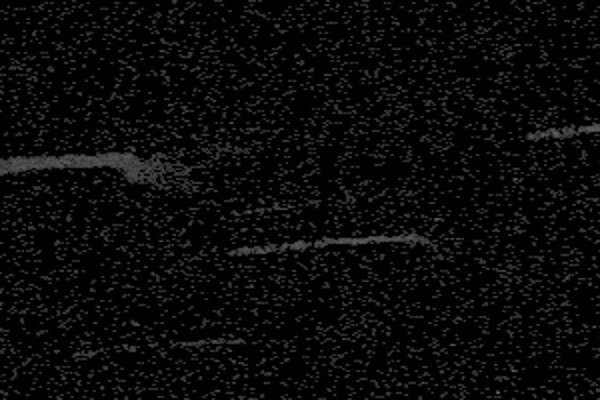}} &
    \fbox{\includegraphics[height=0.45in,width=0.6in]{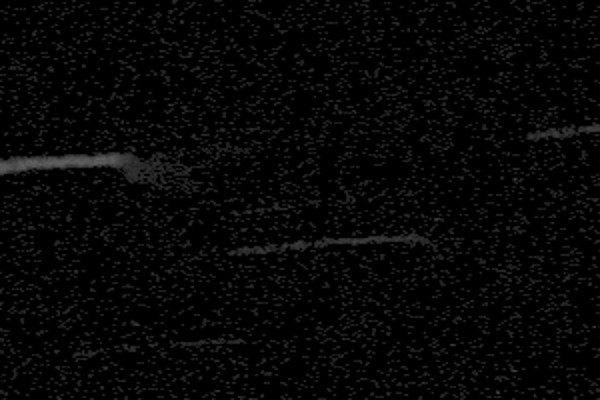}} \\
    & \small\color{gray!90}IETS~\cite{karray_inceptive_2019} & \small\color{gray!90}STCF~\cite{guo_low_2023} & \small\color{gray!90}CrossConv & \small\color{gray!90}MLPF~\cite{guo_low_2023} & \small\color{gray!90}AEDNet~\cite{fang_aednet_2022} & \small\color{gray!90}FEAST~\cite{afshar_event-based_2020} & \small\color{gray!90}FEAST+C~\cite{afshar_event-based_2020}\\
    \tiny\rotatebox{90}{\hspace{.2cm}\color{gray!90}Denoised} & 
    \fbox{\includegraphics[height=0.45in,width=0.6in]{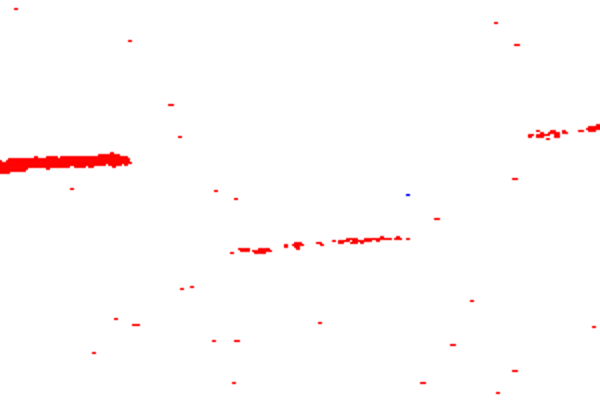}} & 
    \fbox{\includegraphics[height=0.45in,width=0.6in]{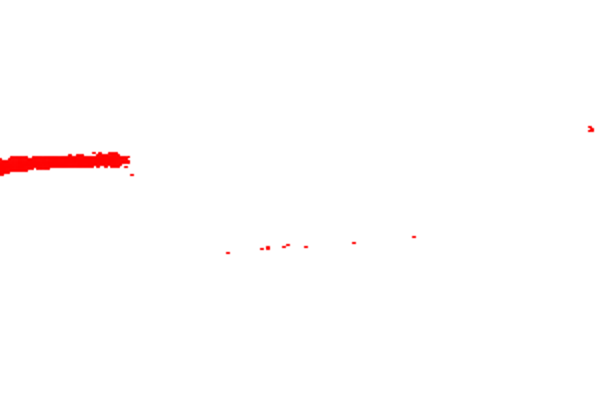}} & 
    \fbox{\includegraphics[height=0.45in,width=0.6in]{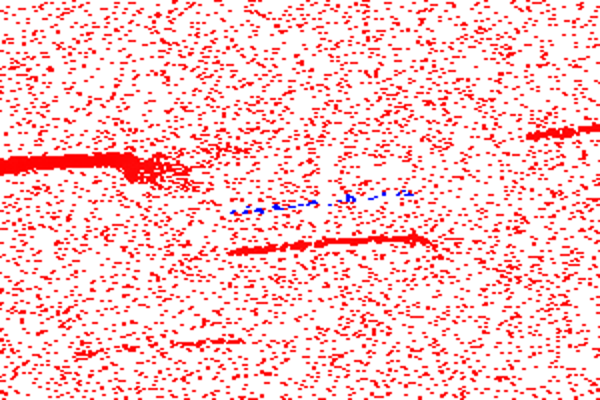}} & 
    \fbox{\includegraphics[height=0.45in,width=0.6in]{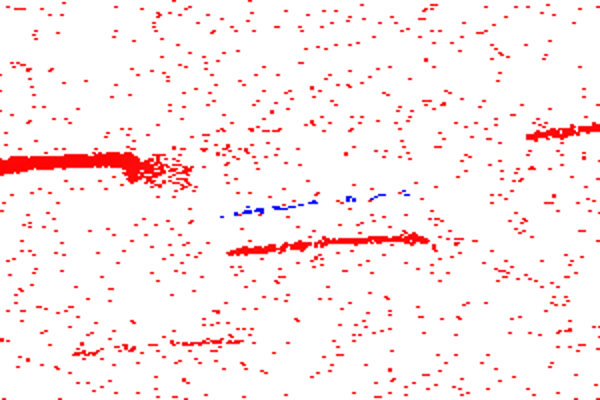}} & 
    \fbox{\includegraphics[height=0.45in,width=0.6in]{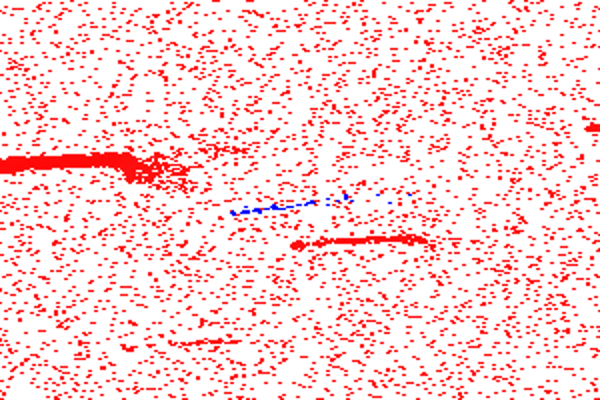}} & 
    \fbox{\includegraphics[height=0.45in,width=0.6in]{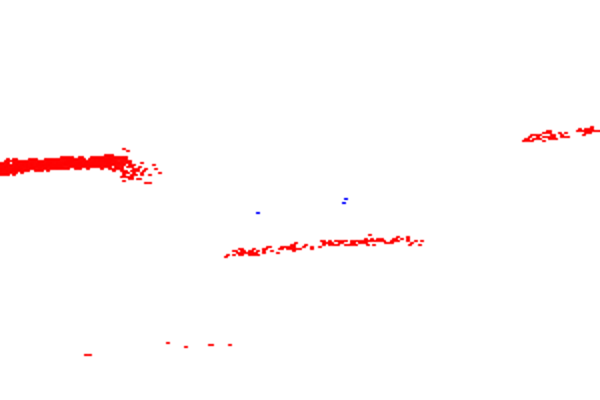}} & 
    \fbox{\includegraphics[height=0.45in,width=0.6in]{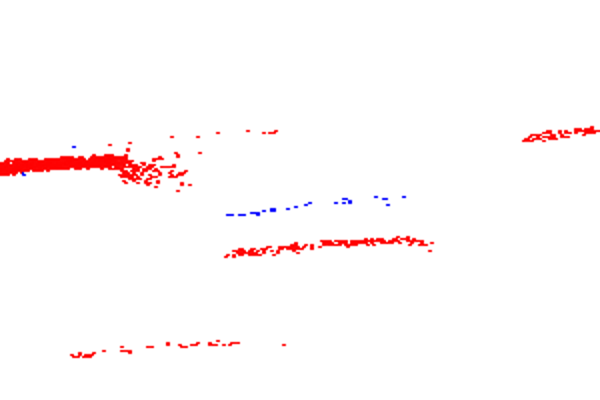}} \\
    \tiny\rotatebox{90}{\color{gray!90}Residual Noise} &
    \fbox{\includegraphics[height=0.45in,width=0.6in]{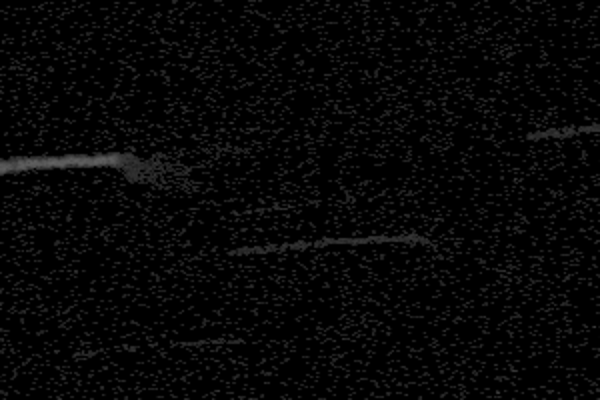}} & 
    \fbox{\includegraphics[height=0.45in,width=0.6in]{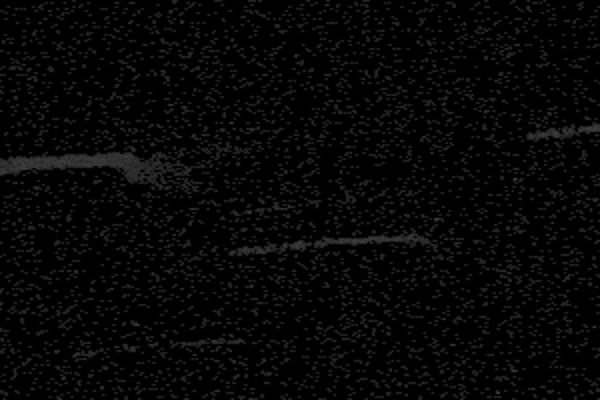}} & 
    \fbox{\includegraphics[height=0.45in,width=0.6in]{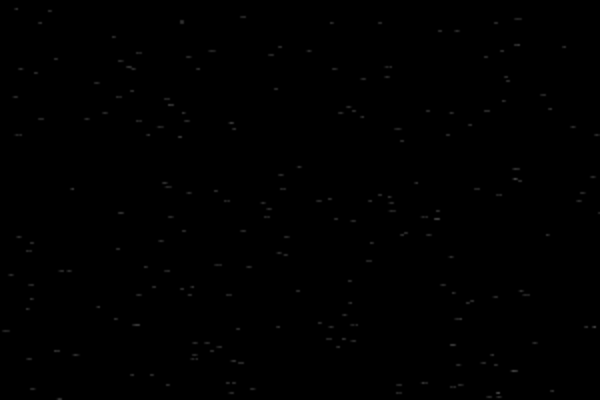}} & 
    \fbox{\includegraphics[height=0.45in,width=0.6in]{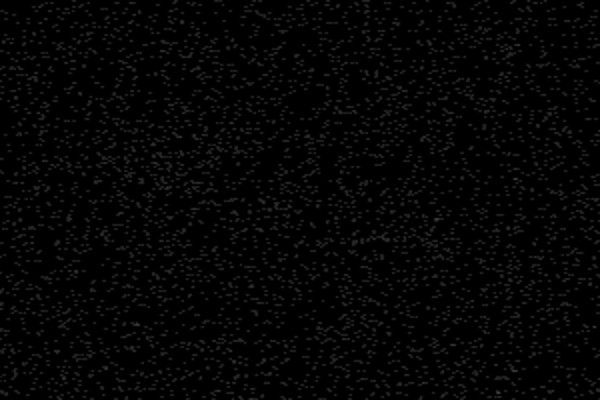}} & 
    \fbox{\includegraphics[height=0.45in,width=0.6in]{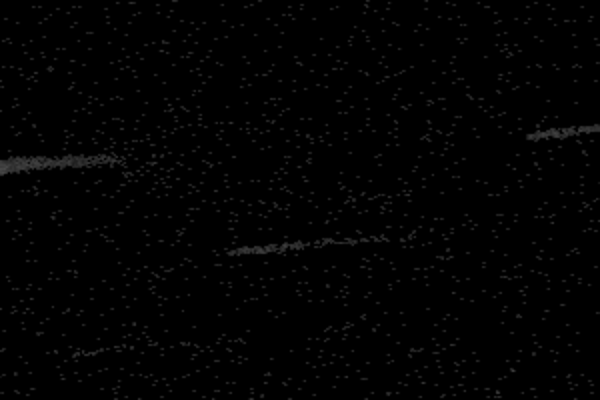}} & 
    \fbox{\includegraphics[height=0.45in,width=0.6in]{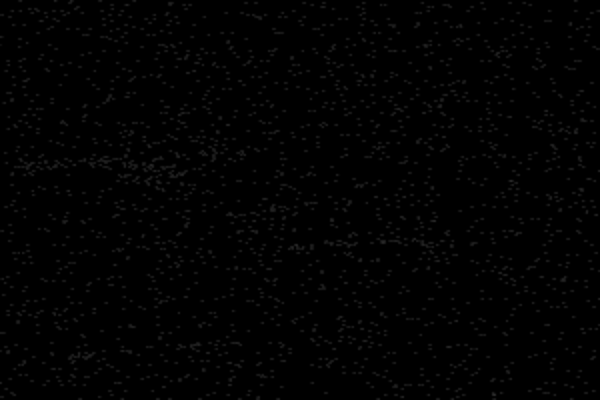}} & 
    \fbox{\includegraphics[height=0.45in,width=0.6in]{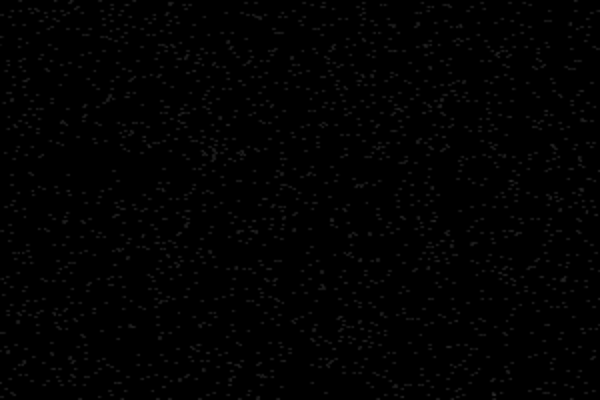}} \\
\end{tabular}
\end{subfigure}

\caption{Visual comparison of noise filtering algorithms on the Ev-Satellite dataset. \textbf{Top row:} Denoised outputs generated by each algorithm. \textbf{Bottom row:} Residual noise events, representing true negatives. Events are colour-coded: blue for satellites and red for stars.FEAST+C is short for FEAST+Classifier.}
\label{fig:qualitativeresults}
\end{figure*}

\textbf{Qualitative Evaluations}. Figure~\ref{fig:qualitativeresults} illustrates the performance of each noise filtering algorithm across three test datasets from the Ev-Satellite dataset. Rows 1 to 4, show multiple satellites moving in the same direction. Rows 5 to 8, show a single satellite with a non-linear motion from the wind. Rows 9 to 12 show a single faint satellite among dense stars.

Among the algorithms, FEAST+classifier consistently outperforms the other algorithms, demonstrating superior noise and hot pixel removal capabilities. In contrast, EvFlow's performance varies across different recordings, indicating inconsistency over multiple recordings. STDF, CrossConv, and AEDAT tend to ignore a higher number of noise events, while IETS and YNoise often remove too many satellite events, resulting in higher false negatives. Overall, FEAST+classifier proves to be the most effective in reducing noise and preserving relevant events.

\begin{figure}[htbp]
\centering
\begin{subfigure}{\linewidth}
   \centering
   \includegraphics[width=\linewidth]{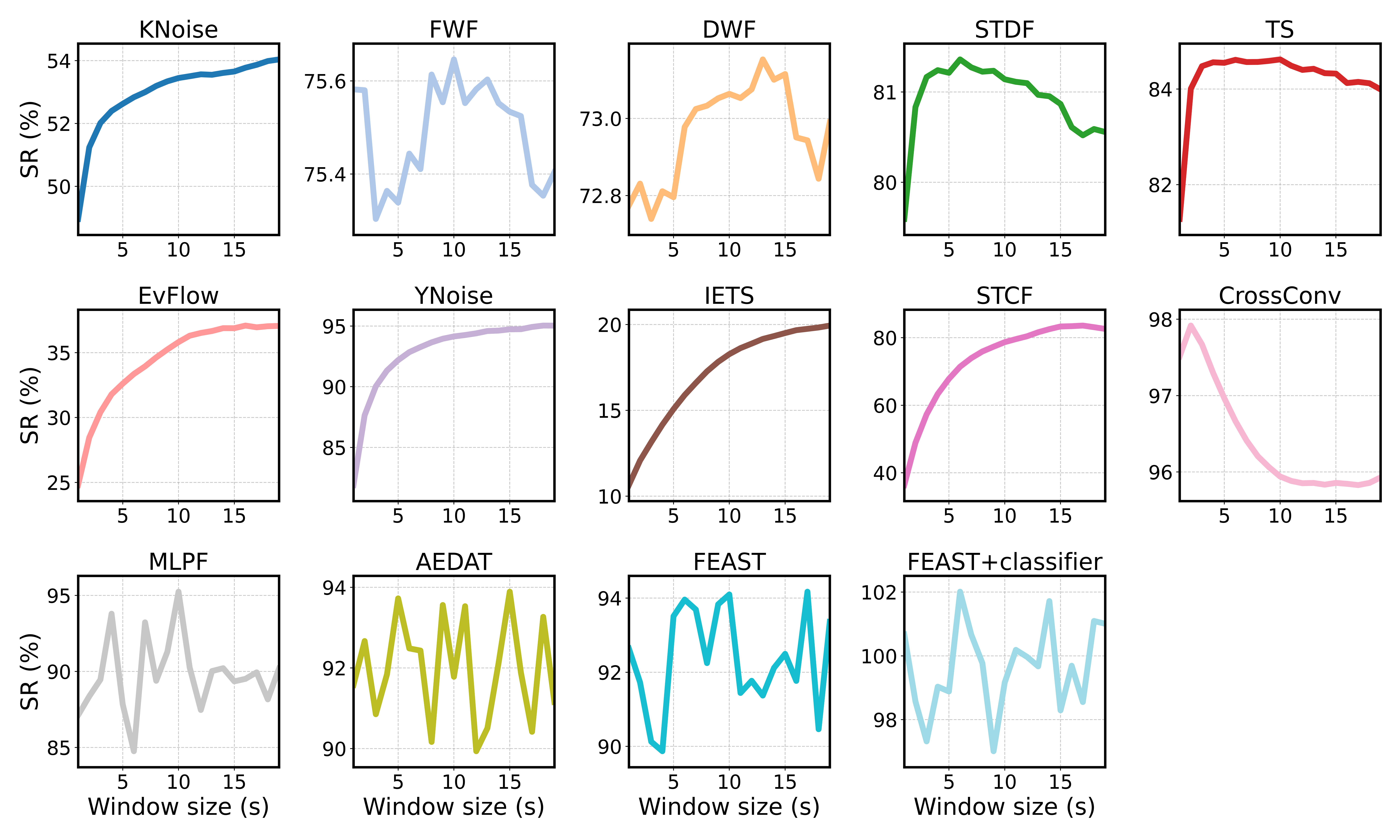}
   \caption{Signal Retain (SR) metric in \%.}
   \label{fig:windowsize_NR}
\end{subfigure}

\vspace{1em}

\begin{subfigure}{\linewidth}
   \centering
   \includegraphics[width=\linewidth]{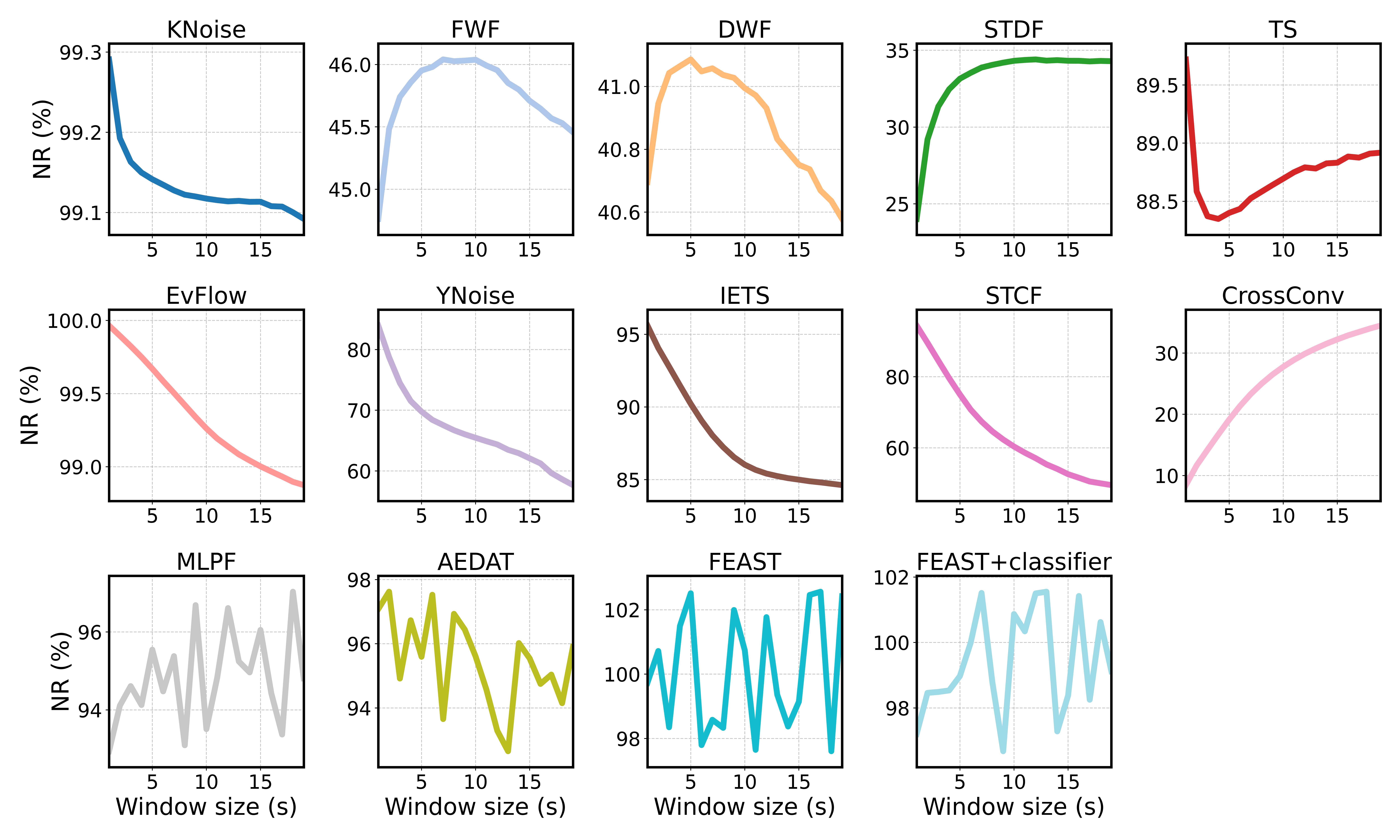}
   \caption{Noise Removal (NR) metric in \%.}
   \label{fig:windowsize_SR}
\end{subfigure}

\caption{Performance comparison of all noise filtering algorithms over window sizes ranging from 1 to 20 seconds. The performance of each algorithm is presented in its own plot because of the significant differences among each methods, and to avoids the need to zoom in on individual curves.}

\label{fig:windowsize}
\end{figure}

\textbf{Performance over Window Size}. We evaluated the performance of various noise filtering algorithms across different window sizes to determine the optimal event window length for accurate noise filtering. Using Receiver Operating Characteristic (ROC) curves, we selected parameters that yielded maximum accuracy while keeping all other parameters constant. We systematically varied the window size from 1 to 20 seconds, incrementing by 1 second at each step. Additionally, we applied a stride such that, for a given window size $w_s=1$ second, the stride length was set to $w_s/5$. This striding was performed from the beginning to the end of the recording, and we calculated the average accuracy for each window size. This procedure was repeated for all window sizes using both the Signal Retain (SR) and Noise Removal (NR) metrics. The results of this experiment are summarized in Figure~\ref{fig:windowsize}.

Our findings suggest that increasing the window size enhances noise filtering performance for logical-based algorithms, likely because larger windows provide more samples for the algorithms to process. This improvement was observed exclusively in the logical-based algorithms. In contrast, we observed that the noise removal performance decreased with larger window sizes for the IETS, STCF, and YNoise algorithms, while their signal retention performance increased. This indicates that these algorithms denoise less effectively with larger windows. On the contrary, the performance of the learning-based algorithm remained largely unchanged and without a clear upward or downward trend when the window size increases.

Among all the algorithms tested, the combination of FEAST with a classifier achieved the best noise removal and signal retention performance. This was followed by the CrossConv algorithm, which demonstrated high signal retention but was less effective in noise removal since it prioritises hot pixels. The TS algorithm emerged as the second-best denoiser overall, providing better noise removal at the expense of removing more signal information.

\section{Discussion}
Filtering the noise in sparse event streams can reduce visual noise but comes at the cost of removing real satellite events. In this work, the best algorithm can only preserve up to 95.98\%  of the satellite events. In very sparse satellite streams, filtering the noise might remove significant portions of actual data, which can be undesirable.

Hot pixels are the most dominant type of noise which artificially increases the event rate, especially in sparse scenes. This makes the task of motion estimation (e.g., CMax) and tracking~\cite{ralph2022real} more challenging, therefore, their removal is more critical than background noise. However, most noise filtering algorithms remove hot pixels in pre- or post-processing or by aggressive noise filtering, which can reduce true positives. We found that KNoise was the most effective at removing hot pixels among the logical-based algorithm, followed by our proposed lightweight CrossConv algorithm. Thus, in very sparse scenes, algorithms that prioritize hot pixel removal are most beneficial.

Noise filtering also assists in other SSA tasks like determining the limiting magnitude. Previous work determined magnitude by merging event streams using a binocular event camera setup~\cite{marcireau2023binocular}. Our study provides insights into noise filtering from a single event stream and serves as a comparison to the binocular approach.

Given the vast data from Astrosite, developing end-to-end, data-driven noise filtering algorithms that exploit structural cues is practical. Such algorithms can account for variations in event streams due to sky brightness, star brightness, satellite glinting or tumbling, and platform vibration. While FEAST-like architectures have been demonstrated on hardware \cite{jose2024fpga,mehrabi2023optimized,mehrabi2023efficient}, hardware-based noise filtering has not yet been implemented. However, due to the event-driven nature of the FEAST algorithm, real-time noise filtering could be achieved through hardware within the event camera's logic circuits in future work.

We performed noise filtering by scanning the sky at a constant speed for over 30 seconds. Alternative methods include scanning at multiple speeds for longer periods or using specialised mirrors or liquid lenses to artificially increase contrast and enable rapid focus adjustments~\cite{ralph2023shake}. The idea is that if an object appears and reappears during focus changes, it is likely a real object (e.g., a satellite). However, these methods may compromise the real-time performance of the system but have the potential to provide more accurate noise-filtering results.

\section{Conclusion}

In this paper, we presented three algorithms that achieve high-accuracy noise filtering on a high-resolution and real-world satellite dataset. Our primary goal was to address the challenge of noise filtering of very sparse satellite signals that can be obscured by excessive noise and hot pixels, demonstrating the best approach to filter noise from the event data while preserving the satellite signals. We conducted extensive experiments using 11 state-of-the-art noise filtering methods and provided a comparative analysis of their performance. We utilized the ROC curve for performance comparison over the entire parameter range to ensure a fair comparison between the algorithms. The experimental results highlight the superiority of the learning-based FEAST algorithm in removing noise and hot pixels. Additionally, the logical-based EvFlow algorithm proved most effective in preserving satellite signals when using appropriate parameters. We hope that our work will contribute valuable insights to the event community and spur future research on event-based sparse noise filtering for SSA applications.

\setlength{\parskip}{10pt}

\textbf{Acknowledgements}. This work was supported by the Air Force Office of Scientific Research (AFOSR) under grant FA2386-23-1-4005.

\bibliographystyle{IEEEtran}
\bibliography{egbib}

\begin{thebibliography}{10}
\providecommand{\url}[1]{#1}
\csname url@samestyle\endcsname
\providecommand{\newblock}{\relax}
\providecommand{\bibinfo}[2]{#2}
\providecommand{\BIBentrySTDinterwordspacing}{\spaceskip=0pt\relax}
\providecommand{\BIBentryALTinterwordstretchfactor}{4}
\providecommand{\BIBentryALTinterwordspacing}{\spaceskip=\fontdimen2\font plus
\BIBentryALTinterwordstretchfactor\fontdimen3\font minus \fontdimen4\font\relax}
\providecommand{\BIBforeignlanguage}[2]{{%
\expandafter\ifx\csname l@#1\endcsname\relax
\typeout{** WARNING: IEEEtran.bst: No hyphenation pattern has been}%
\typeout{** loaded for the language `#1'. Using the pattern for}%
\typeout{** the default language instead.}%
\else
\language=\csname l@#1\endcsname
\fi
#2}}
\providecommand{\BIBdecl}{\relax}
\BIBdecl

\bibitem{lichtsteiner_128times128_2008}
\BIBentryALTinterwordspacing
P.~Lichtsteiner, C.~Posch, and T.~Delbruck, ``A 128\${\textbackslash}times\$128 120 {dB} 15 \${\textbackslash}mu\$s {Latency} {Asynchronous} {Temporal} {Contrast} {Vision} {Sensor},'' \emph{IEEE Journal of Solid-State Circuits}, vol.~43, no.~2, pp. 566--576, 2008. [Online]. Available: \url{http://ieeexplore.ieee.org/document/4444573/}
\BIBentrySTDinterwordspacing

\bibitem{Finateu2020510A1}
T.~Finateu, A.~Niwa, D.~Matolin, K.~Tsuchimoto, A.~Mascheroni, E.~Reynaud, P.~Mostafalu, F.~T. Brady, L.~Chotard, F.~Legoff, H.~Takahashi, H.~Wakabayashi, Y.~Oike, and C.~Posch, ``5.10 a 1280×720 back-illuminated stacked temporal contrast event-based vision sensor with 4.86µm pixels, 1.066geps readout, programmable event-rate controller and compressive data-formatting pipeline,'' \emph{2020 IEEE International Solid- State Circuits Conference - (ISSCC)}, pp. 112--114, 2020.

\bibitem{afshar_event-based_ssa}
\BIBentryALTinterwordspacing
S.~Afshar, A.~P. Nicholson, A.~Van~Schaik, and G.~Cohen, ``Event-{Based} {Object} {Detection} and {Tracking} for {Space} {Situational} {Awareness},'' \emph{IEEE Sensors J.}, vol.~20, no.~24, pp. 15\,117--15\,132, Dec. 2020. [Online]. Available: \url{https://ieeexplore.ieee.org/document/9142352/}
\BIBentrySTDinterwordspacing

\bibitem{graca_optimal_2023}
\BIBentryALTinterwordspacing
R.~Graca, B.~McReynolds, and T.~Delbruck, ``Optimal biasing and physical limits of {DVS} event noise,'' Apr. 2023, arXiv:2304.04019 [cs, eess]. [Online]. Available: \url{http://arxiv.org/abs/2304.04019}
\BIBentrySTDinterwordspacing

\bibitem{Graca_2023_CVPR}
R.~Gra\c{c}a, B.~McReynolds, and T.~Delbruck, ``Shining light on the dvs pixel: A tutorial and discussion about biasing and optimization,'' in \emph{Proceedings of the IEEE/CVF Conference on Computer Vision and Pattern Recognition (CVPR) Workshops}, June 2023, pp. 4045--4053.

\bibitem{Mcreynolds2023ExploitingAD}
\BIBentryALTinterwordspacing
B.~Mcreynolds, R.~Graça, and T.~Delbruck, ``Exploiting alternating dvs shot noise event pair statistics to reduce background activity,'' 2023. [Online]. Available: \url{https://api.semanticscholar.org/CorpusID:258041241}
\BIBentrySTDinterwordspacing

\bibitem{nozaki_temperature_2017}
\BIBentryALTinterwordspacing
Y.~Nozaki and T.~Delbruck, ``Temperature and {Parasitic} {Photocurrent} {Effects} in {Dynamic} {Vision} {Sensors},'' \emph{IEEE Trans. Electron Devices}, vol.~64, no.~8, pp. 3239--3245, Aug. 2017. [Online]. Available: \url{https://ieeexplore.ieee.org/document/7962235/}
\BIBentrySTDinterwordspacing

\bibitem{graca_unraveling_2021}
\BIBentryALTinterwordspacing
R.~Graca and T.~Delbruck, ``Unraveling the paradox of intensity-dependent {DVS} pixel noise,'' Sep. 2021, arXiv:2109.08640 [physics]. [Online]. Available: \url{http://arxiv.org/abs/2109.08640}
\BIBentrySTDinterwordspacing

\bibitem{Hu_2021_CVPR}
Y.~Hu, S.-C. Liu, and T.~Delbruck, ``v2e: From video frames to realistic dvs events,'' in \emph{Proceedings of the IEEE/CVF Conference on Computer Vision and Pattern Recognition (CVPR) Workshops}, June 2021, pp. 1312--1321.

\bibitem{guo_low_2023}
\BIBentryALTinterwordspacing
S.~Guo and T.~Delbruck, ``Low {Cost} and {Latency} {Event} {Camera} {Background} {Activity} {Denoising},'' \emph{IEEE Trans. Pattern Anal. Mach. Intell.}, vol.~45, no.~1, pp. 785--795, Jan. 2023. [Online]. Available: \url{https://ieeexplore.ieee.org/document/9720086/}
\BIBentrySTDinterwordspacing

\bibitem{gallego_unifying_2018}
\BIBentryALTinterwordspacing
G.~Gallego, H.~Rebecq, and D.~Scaramuzza, ``A unifying contrast maximization framework for event cameras, with applications to motion, depth, and optical flow estimation,'' in \emph{2018 {IEEE}/{CVF} Conference on Computer Vision and Pattern Recognition}.\hskip 1em plus 0.5em minus 0.4em\relax {IEEE}, pp. 3867--3876. [Online]. Available: \url{https://ieeexplore.ieee.org/document/8578505/}
\BIBentrySTDinterwordspacing

\bibitem{marcireau2023binocular}
A.~Marcireau, S.~Afshar, N.~Ralph, I.~Jones, and G.~Cohen, ``Binocular telescope for neuromorphic space situational awareness,'' in \emph{Proceedings of the Advanced Maui Optical and Space Surveillance (AMOS) Technologies Conference}, 2023, p. 137.

\bibitem{ralph_astrometric_2023}
\BIBentryALTinterwordspacing
N.~O. Ralph, A.~Marcireau, S.~Afshar, N.~Tothill, A.~Van~Schaik, and G.~Cohen, ``\BIBforeignlanguage{en}{Astrometric calibration and source characterisation of the latest generation neuromorphic event-based cameras for space imaging},'' \emph{\BIBforeignlanguage{en}{Astrodyn}}, vol.~7, no.~4, pp. 415--443, Dec. 2023. [Online]. Available: \url{https://link.springer.com/10.1007/s42064-023-0168-2}
\BIBentrySTDinterwordspacing

\bibitem{mcreynolds_demystifying_2023}
\BIBentryALTinterwordspacing
B.~McReynolds, R.~Graca, R.~Oliver, M.~Nishiguchi, and T.~Delbruck, ``Demystifying {Event}-based {Sensor} {Biasing} to {Optimize} {Signal} to {Noise} for {Space} {Domain} {Awareness},'' Sep. 2023, publisher: s.n. [Online]. Available: \url{https://www.zora.uzh.ch/id/eprint/254194}
\BIBentrySTDinterwordspacing

\bibitem{barrios2018less}
J.~Barrios-Avil{\'e}s, A.~Rosado-Mu{\~n}oz, L.~D. Medus, M.~Bataller-Mompe{\'a}n, and J.~F. Guerrero-Mart{\'\i}nez, ``Less data same information for event-based sensors: A bioinspired filtering and data reduction algorithm,'' \emph{Sensors}, vol.~18, no.~12, p. 4122, 2018.

\bibitem{khodamoradi_on-space_2018}
\BIBentryALTinterwordspacing
A.~Khodamoradi and R.~Kastner, ``O({N})-{Space} {Spatiotemporal} {Filter} for {Reducing} {Noise} in {Neuromorphic} {Vision} {Sensors},'' \emph{IEEE Trans. Emerg. Topics Comput.}, pp. 1--1, 2018. [Online]. Available: \url{http://ieeexplore.ieee.org/document/8244294/}
\BIBentrySTDinterwordspacing

\bibitem{liu2015design}
H.~Liu, C.~Brandli, C.~Li, S.-C. Liu, and T.~Delbruck, ``Design of a spatiotemporal correlation filter for event-based sensors,'' in \emph{2015 IEEE International Symposium on Circuits and Systems (ISCAS)}.\hskip 1em plus 0.5em minus 0.4em\relax IEEE, 2015, pp. 722--725.

\bibitem{benosman2013event}
R.~Benosman, C.~Clercq, X.~Lagorce, S.-H. Ieng, and C.~Bartolozzi, ``Event-based visual flow,'' \emph{IEEE transactions on neural networks and learning systems}, vol.~25, no.~2, pp. 407--417, 2013.

\bibitem{delbruck_frame-free_2008}
\BIBentryALTinterwordspacing
T.~Delbruck, ``Frame-free dynamic digital vision,'' Mar. 2008, publisher: University of Tokyo. [Online]. Available: \url{https://www.zora.uzh.ch/id/eprint/17620}
\BIBentrySTDinterwordspacing

\bibitem{baldwin_edncnn}
R.~W. Baldwin, M.~Almatrafi, V.~Asari, and K.~Hirakawa, ``Event probability mask (epm) and event denoising convolutional neural network (edncnn) for neuromorphic cameras,'' in \emph{Proceedings of the IEEE/CVF Conference on Computer Vision and Pattern Recognition (CVPR)}, June 2020.

\bibitem{karray_inceptive_2019}
\BIBentryALTinterwordspacing
R.~W. Baldwin, M.~Almatrafi, J.~R. Kaufman, V.~Asari, and K.~Hirakawa, ``\BIBforeignlanguage{en}{Inceptive {Event} {Time}-{Surfaces} for {Object} {Classification} {Using} {Neuromorphic} {Cameras}},'' in \emph{\BIBforeignlanguage{en}{Image {Analysis} and {Recognition}}}, F.~Karray, A.~Campilho, and A.~Yu, Eds.\hskip 1em plus 0.5em minus 0.4em\relax Cham: Springer International Publishing, 2019, vol. 11663, pp. 395--403, series Title: Lecture Notes in Computer Science. [Online]. Available: \url{http://link.springer.com/10.1007/978-3-030-27272-2_35}
\BIBentrySTDinterwordspacing

\bibitem{wang2020joint}
Z.~W. Wang, P.~Duan, O.~Cossairt, A.~Katsaggelos, T.~Huang, and B.~Shi, ``Joint filtering of intensity images and neuromorphic events for high-resolution noise-robust imaging,'' in \emph{Proceedings of the IEEE/CVF Conference on Computer Vision and Pattern Recognition}, 2020, pp. 1609--1619.

\bibitem{fang_aednet_2022}
\BIBentryALTinterwordspacing
H.~Fang, J.~Wu, L.~Li, J.~Hou, W.~Dong, and G.~Shi, ``\BIBforeignlanguage{en}{{AEDNet}: {Asynchronous} {Event} {Denoising} with {Spatial}-{Temporal} {Correlation} among {Irregular} {Data}},'' in \emph{\BIBforeignlanguage{en}{Proceedings of the 30th {ACM} {International} {Conference} on {Multimedia}}}.\hskip 1em plus 0.5em minus 0.4em\relax Lisboa Portugal: ACM, Oct. 2022, pp. 1427--1435. [Online]. Available: \url{https://dl.acm.org/doi/10.1145/3503161.3548048}
\BIBentrySTDinterwordspacing

\bibitem{duan_led_2024}
\BIBentryALTinterwordspacing
Y.~Duan, S.~Peng, L.~Zhu, W.~Zhang, Y.~Chang, S.~Zhong, and L.~Yan, ``{LED}: {A} {Large}-scale {Real}-world {Paired} {Dataset} for {Event} {Camera} {Denoising},'' May 2024, arXiv:2405.19718 [cs]. [Online]. Available: \url{http://arxiv.org/abs/2405.19718}
\BIBentrySTDinterwordspacing

\bibitem{afshar_event-based_2020}
\BIBentryALTinterwordspacing
S.~Afshar, N.~Ralph, Y.~Xu, J.~Tapson, A.~v. Schaik, and G.~Cohen, ``\BIBforeignlanguage{en}{Event-{Based} {Feature} {Extraction} {Using} {Adaptive} {Selection} {Thresholds}},'' \emph{\BIBforeignlanguage{en}{Sensors}}, vol.~20, no.~6, p. 1600, Mar. 2020. [Online]. Available: \url{https://www.mdpi.com/1424-8220/20/6/1600}
\BIBentrySTDinterwordspacing

\bibitem{Delbruck_2021_CVPR}
T.~Delbruck, R.~Graca, and M.~Paluch, ``Feedback control of event cameras,'' in \emph{Proceedings of the IEEE/CVF Conference on Computer Vision and Pattern Recognition (CVPR) Workshops}, June 2021, pp. 1324--1332.

\bibitem{alkendi_neuromorphic_2024}
\BIBentryALTinterwordspacing
Y.~Alkendi, R.~Azzam, A.~Ayyad, S.~Javed, L.~Seneviratne, and Y.~Zweiri, ``Neuromorphic {Camera} {Denoising} {Using} {Graph} {Neural} {Network}-{Driven} {Transformers},'' \emph{IEEE Trans. Neural Netw. Learning Syst.}, vol.~35, no.~3, pp. 4110--4124, Mar. 2024. [Online]. Available: \url{https://ieeexplore.ieee.org/document/9893571/}
\BIBentrySTDinterwordspacing

\bibitem{ding_e-mlb_2024}
\BIBentryALTinterwordspacing
S.~Ding, J.~Chen, Y.~Wang, Y.~Kang, W.~Song, J.~Cheng, and Y.~Cao, ``E-{MLB}: {Multilevel} {Benchmark} for {Event}-{Based} {Camera} {Denoising},'' \emph{IEEE Trans. Multimedia}, vol.~26, pp. 65--76, 2024. [Online]. Available: \url{https://ieeexplore.ieee.org/document/10078400/}
\BIBentrySTDinterwordspacing

\bibitem{padala_noise_2018}
\BIBentryALTinterwordspacing
V.~Padala, A.~Basu, and G.~Orchard, ``A {Noise} {Filtering} {Algorithm} for {Event}-{Based} {Asynchronous} {Change} {Detection} {Image} {Sensors} on {TrueNorth} and {Its} {Implementation} on {TrueNorth},'' \emph{Front. Neurosci.}, vol.~12, p. 118, Mar. 2018. [Online]. Available: \url{http://journal.frontiersin.org/article/10.3389/fnins.2018.00118/full}
\BIBentrySTDinterwordspacing

\bibitem{wu_probabilistic_2021}
\BIBentryALTinterwordspacing
J.~Wu, C.~Ma, L.~Li, W.~Dong, and G.~Shi, ``Probabilistic {Undirected} {Graph} {Based} {Denoising} {Method} for {Dynamic} {Vision} {Sensor},'' \emph{IEEE Trans. Multimedia}, vol.~23, pp. 1148--1159, 2021. [Online]. Available: \url{https://ieeexplore.ieee.org/document/9091226/}
\BIBentrySTDinterwordspacing

\bibitem{feng_event_2020}
\BIBentryALTinterwordspacing
Y.~Feng, H.~Lv, H.~Liu, Y.~Zhang, Y.~Xiao, and C.~Han, ``\BIBforeignlanguage{en}{Event {Density} {Based} {Denoising} {Method} for {Dynamic} {Vision} {Sensor}},'' \emph{\BIBforeignlanguage{en}{Applied Sciences}}, vol.~10, no.~6, p. 2024, Mar. 2020. [Online]. Available: \url{https://www.mdpi.com/2076-3417/10/6/2024}
\BIBentrySTDinterwordspacing

\bibitem{lagorce_hots_2017}
\BIBentryALTinterwordspacing
X.~Lagorce, G.~Orchard, F.~Galluppi, B.~E. Shi, and R.~B. Benosman, ``{HOTS}: {A} {Hierarchy} of {Event}-{Based} {Time}-{Surfaces} for {Pattern} {Recognition},'' \emph{IEEE Trans. Pattern Anal. Mach. Intell.}, vol.~39, no.~7, pp. 1346--1359, Jul. 2017. [Online]. Available: \url{http://ieeexplore.ieee.org/document/7508476/}
\BIBentrySTDinterwordspacing

\bibitem{wang_ev-gait_2019}
\BIBentryALTinterwordspacing
Y.~Wang, B.~Du, Y.~Shen, K.~Wu, G.~Zhao, J.~Sun, and H.~Wen, ``{EV}-{Gait}: {Event}-{Based} {Robust} {Gait} {Recognition} {Using} {Dynamic} {Vision} {Sensors},'' in \emph{2019 {IEEE}/{CVF} {Conference} on {Computer} {Vision} and {Pattern} {Recognition} ({CVPR})}.\hskip 1em plus 0.5em minus 0.4em\relax Long Beach, CA, USA: IEEE, Jun. 2019, pp. 6351--6360. [Online]. Available: \url{https://ieeexplore.ieee.org/document/8953966/}
\BIBentrySTDinterwordspacing

\bibitem{guo2020hashheat}
S.~Guo, Z.~Kang, L.~Wang, S.~Li, and W.~Xu, ``Hashheat: An o (c) complexity hashing-based filter for dynamic vision sensor,'' in \emph{2020 25th Asia and South Pacific Design Automation Conference (ASP-DAC)}.\hskip 1em plus 0.5em minus 0.4em\relax IEEE, 2020, pp. 452--457.

\bibitem{acharya2019ebbiot}
J.~Acharya, A.~U. Caycedo, V.~R. Padala, R.~R.~S. Sidhu, G.~Orchard, B.~Ramesh, and A.~Basu, ``Ebbiot: A low-complexity tracking algorithm for surveillance in iovt using stationary neuromorphic vision sensors,'' in \emph{2019 32nd IEEE International System-on-Chip Conference (SOCC)}.\hskip 1em plus 0.5em minus 0.4em\relax IEEE, 2019, pp. 318--323.

\bibitem{mohan2020ebbinnot}
V.~Mohan, D.~Singla, T.~Pulluri, A.~Ussa, P.~K. Gopalakrishnan, P.-S. Sun, B.~Ramesh, and A.~Basu, ``Ebbinnot: A hardware efficient hybrid event-frame tracker for stationary dynamic vision sensors,'' \emph{arXiv preprint arXiv:2006.00422}, 2020.

\bibitem{cladera2020device}
F.~Cladera, A.~Bisulco, D.~Kepple, V.~Isler, and D.~D. Lee, ``On-device event filtering with binary neural networks for pedestrian detection using neuromorphic vision sensors,'' in \emph{2020 IEEE International Conference on Image Processing (ICIP)}.\hskip 1em plus 0.5em minus 0.4em\relax IEEE, 2020, pp. 3084--3088.

\bibitem{bose202151}
S.~K. Bose, D.~Singla, and A.~Basu, ``A 51.3-tops/w, 134.4-gops in-memory binary image filtering in 65-nm cmos,'' \emph{IEEE Journal of Solid-State Circuits}, vol.~57, no.~1, pp. 323--335, 2021.

\bibitem{Duan_2021_CVPR}
P.~Duan, Z.~W. Wang, X.~Zhou, Y.~Ma, and B.~Shi, ``Eventzoom: Learning to denoise and super resolve neuromorphic events,'' in \emph{Proceedings of the IEEE/CVF Conference on Computer Vision and Pattern Recognition (CVPR)}, June 2021, pp. 12\,824--12\,833.

\bibitem{linares2019low}
A.~Linares-Barranco, F.~Perez-Pe{\~n}a, D.~P. Moeys, F.~Gomez-Rodriguez, G.~Jimenez-Moreno, S.-C. Liu, and T.~Delbruck, ``Low latency event-based filtering and feature extraction for dynamic vision sensors in real-time fpga applications,'' \emph{IEEE Access}, vol.~7, pp. 134\,926--134\,942, 2019.

\bibitem{duan_guided_2021}
\BIBentryALTinterwordspacing
P.~Duan, Z.~Wang, B.~Shi, O.~Cossairt, T.~Huang, and A.~Katsaggelos, ``Guided {Event} {Filtering}: {Synergy} between {Intensity} {Images} and {Neuromorphic} {Events} for {High} {Performance} {Imaging},'' \emph{IEEE Trans. Pattern Anal. Mach. Intell.}, pp. 1--1, 2021. [Online]. Available: \url{https://ieeexplore.ieee.org/document/9541050/}
\BIBentrySTDinterwordspacing

\bibitem{cohen2018approaches}
G.~Cohen, S.~Afshar, and A.~Van~Schaik, ``Approaches for astrometry using event-based sensors,'' in \emph{Advanced Maui Optical and Space Surveillance (AMOS) Technologies Conference}, 2018, p.~25.

\bibitem{cohen2019event}
G.~Cohen, S.~Afshar, B.~Morreale, T.~Bessell, A.~Wabnitz, M.~Rutten, and A.~van Schaik, ``Event-based sensing for space situational awareness,'' \emph{The Journal of the Astronautical Sciences}, vol.~66, pp. 125--141, 2019.

\bibitem{mcmahon2021commercial}
P.~N. McMahon-Crabtree and D.~G. Monet, ``Commercial-off-the-shelf event-based cameras for space surveillance applications,'' \emph{Applied Optics}, vol.~60, no.~25, pp. G144--G153, 2021.

\bibitem{zolnowski2019observational}
M.~Zo{\l}nowski, R.~Reszelewski, D.~P. Moeys, T.~Delbruck, and K.~Kami{\'n}ski, ``Observational evaluation of event cameras performance in optical space surveillance,'' in \emph{NEO and Debris Detection Conference, Darmstadt, Germany}, 2019.

\bibitem{ralph2023astrometric}
N.~O. Ralph, A.~Marcireau, S.~Afshar, N.~Tothill, A.~Van~Schaik, and G.~Cohen, ``Astrometric calibration and source characterisation of the latest generation neuromorphic event-based cameras for space imaging,'' \emph{Astrodynamics}, vol.~7, no.~4, pp. 415--443, 2023.

\bibitem{ralph2023shake}
N.~Ralph, D.~Maybour, A.~Marcireau, I.~Jones, A.~De~Horta, and G.~Cohen, ``Shake before use: Artificial contrast generation for improved space imaging using neuromorphic event-based vision sensors,'' in \emph{Proceedings of the Advanced Maui Optical and Space Surveillance (AMOS) Technologies Conference}, 2023, p. 161.

\bibitem{ralph2022real}
N.~Ralph, D.~Joubert, A.~Jolley, S.~Afshar, N.~Tothill, A.~Van~Schaik, and G.~Cohen, ``Real-time event-based unsupervised feature consolidation and tracking for space situational awareness,'' \emph{Frontiers in neuroscience}, vol.~16, p. 821157, 2022.

\bibitem{oliver2022event}
R.~Oliver, B.~McReynolds, and D.~Savransky, ``Event-based sensor multiple hypothesis tracker for space domain awareness,'' in \emph{Advanced Maui Optical and Space Surveillance Technologies Conference (AMOS)}.\hskip 1em plus 0.5em minus 0.4em\relax University of Zurich, 2022.

\bibitem{ralph2023exploring}
N.~O. Ralph, ``Exploring space situational awareness using neuromorphic event-based cameras,'' 2023.

\bibitem{jolley2023neuromorphic}
A.~Jolley, S.~Afshar, G.~Cohen, R.~Lazarus~Pahlavani, and A.~Lambert, ``Neuromorphic sensor event-rate monitoring for satellite characterization,'' \emph{Journal of Spacecraft and Rockets}, vol.~60, no.~3, pp. 753--764, 2023.

\bibitem{jolley2022evaluation}
A.~Jolley, G.~Cohen, D.~Joubert, and A.~Lambert, ``Evaluation of event-based sensors for satellite material characterization,'' \emph{Journal of Spacecraft and Rockets}, vol.~59, no.~2, pp. 627--636, 2022.

\bibitem{jolley2022characterising}
A.~Jolley and G.~Cohen, ``Characterising satellites using neuromorphic sensor multicolour broadband event-rates,'' \emph{44th COSPAR Scientific Assembly. Held 16-24 July}, vol.~44, p. 3162, 2022.

\bibitem{jolley2019use}
A.~Jolley, G.~Cohen, and A.~Lambert, ``Use of neuromorphic sensors for satellite material characterisation,'' in \emph{Imaging Systems and Applications}.\hskip 1em plus 0.5em minus 0.4em\relax Optica Publishing Group, 2019, pp. IM1B--4.

\bibitem{Oliver2024}
\BIBentryALTinterwordspacing
R.~Oliver, ``\BIBforeignlanguage{English}{An event-based vision sensor simulation framework for space domain awareness applications},'' Ph.D. dissertation, ProQuest Dissertations and Theses, 2024, copyright - Database copyright ProQuest LLC; ProQuest does not claim copyright in the individual underlying works; Last updated - 2024-09-05. [Online]. Available: \url{http://ezproxy.uws.edu.au/login?url=https://www.proquest.com/dissertations-theses/event-based-vision-sensor-simulation-framework/docview/3100813755/se-2}
\BIBentrySTDinterwordspacing

\bibitem{gallego2019focus}
G.~Gallego, M.~Gehrig, and D.~Scaramuzza, ``Focus is all you need: Loss functions for event-based vision,'' in \emph{Proceedings of the IEEE/CVF Conference on Computer Vision and Pattern Recognition}, 2019, pp. 12\,280--12\,289.

\bibitem{lang2010astrometry}
D.~Lang, D.~W. Hogg, K.~Mierle, M.~Blanton, and S.~Roweis, ``Astrometry. net: Blind astrometric calibration of arbitrary astronomical images,'' \emph{The astronomical journal}, vol. 139, no.~5, p. 1782, 2010.

\bibitem{andrae2023gaia}
R.~Andrae, M.~Fouesneau, R.~Sordo, C.~Bailer-Jones, T.~Dharmawardena, J.~Rybizki, F.~De~Angeli, H.~Lindstr{\o}m, D.~Marshall, R.~Drimmel \emph{et~al.}, ``Gaia data release 3-analysis of the gaia bp/rp spectra using the general stellar parameterizer from photometry,'' \emph{Astronomy \& Astrophysics}, vol. 674, p. A27, 2023.

\bibitem{vallado2012two}
D.~A. Vallado and P.~J. Cefola, ``Two-line element sets--practice and use,'' in \emph{63rd International Astronautical Congress, Naples, Italy}, 2012, pp. 1--14.

\bibitem{bethi_optimized_2022}
\BIBentryALTinterwordspacing
Y.~Bethi, Y.~Xu, G.~Cohen, A.~Van~Schaik, and S.~Afshar, ``An {Optimized} {Deep} {Spiking} {Neural} {Network} {Architecture} {Without} {Gradients},'' \emph{IEEE Access}, vol.~10, pp. 97\,912--97\,929, 2022. [Online]. Available: \url{https://ieeexplore.ieee.org/document/9864144/}
\BIBentrySTDinterwordspacing

\bibitem{tapson2013synthesis}
J.~C. Tapson, G.~K. Cohen, S.~Afshar, K.~M. Stiefel, Y.~Buskila, R.~M. Wang, T.~J. Hamilton, and A.~van Schaik, ``Synthesis of neural networks for spatio-temporal spike pattern recognition and processing,'' \emph{Frontiers in neuroscience}, vol.~7, p. 153, 2013.

\bibitem{el2022neuromorphic}
S.~El~Arja, ``Neuromorphic perception for greenhouse technology using event-based sensors,'' 2022.

\bibitem{cohen2018spatial}
G.~Cohen, S.~Afshar, G.~Orchard, J.~Tapson, R.~Benosman, and A.~van Schaik, ``Spatial and temporal downsampling in event-based visual classification,'' \emph{IEEE Transactions on Neural Networks and Learning Systems}, vol.~29, no.~10, pp. 5030--5044, 2018.

\bibitem{van_schaik_online_2015}
\BIBentryALTinterwordspacing
A.~Van~Schaik and J.~Tapson, ``\BIBforeignlanguage{en}{Online and adaptive pseudoinverse solutions for {ELM} weights},'' \emph{\BIBforeignlanguage{en}{Neurocomputing}}, vol. 149, pp. 233--238, Feb. 2015. [Online]. Available: \url{https://linkinghub.elsevier.com/retrieve/pii/S0925231214011485}
\BIBentrySTDinterwordspacing

\bibitem{Navaro2023-mlpf-dvs-denoising}
A.~R. Navaro, S.~Guo, A.~Gnaneswaran, K.~Vijayakumar, A.~L. Barranco, T.~Aarrestad, R.~Kastner, and T.~Delbruck, ``{Within-Camera} multilayer perceptron {DVS} denoising,'' in \emph{2023 {IEEE/CVF} Conference on Computer Vision and Pattern Recognition Workshops ({CVPRW})}, 2023.

\bibitem{qi2017pointnet}
C.~R. Qi, H.~Su, K.~Mo, and L.~J. Guibas, ``Pointnet: Deep learning on point sets for 3d classification and segmentation,'' in \emph{Proceedings of the IEEE conference on computer vision and pattern recognition}, 2017, pp. 652--660.

\bibitem{jose2024fpga}
P.~C. Jose, Y.~Xu, A.~Van~Schaik, and R.~Wang, ``An fpga implementation of an event-driven unsupervised feature extraction algorithm for pattern recognition,'' in \emph{2024 IEEE International Symposium on Circuits and Systems (ISCAS)}.\hskip 1em plus 0.5em minus 0.4em\relax IEEE, 2024, pp. 1--5.

\bibitem{mehrabi2023optimized}
A.~Mehrabi, Y.~Bethi, A.~van Schaik, and S.~Afshar, ``An optimized multi-layer spiking neural network implementation in fpga without multipliers,'' \emph{Procedia Computer Science}, vol. 222, pp. 407--414, 2023.

\bibitem{mehrabi2023efficient}
A.~Mehrabi, Y.~Bethi, A.~van Schaik, A.~Wabnitz, and S.~Afshar, ``Efficient implementation of a multi-layer gradient-free online-trainable spiking neural network on fpga,'' \emph{arXiv preprint arXiv:2305.19468}, 2023.

\end{thebibliography}

\vspace{11pt}

\vfill

\end{document}